\documentclass[11pt]{article}

\usepackage[final]{acl}

\usepackage{times}
\usepackage{latexsym}

\usepackage[T1]{fontenc}

\usepackage[utf8]{inputenc}

\usepackage{microtype}

\usepackage{inconsolata}

\usepackage{graphicx}
\usepackage{CJKutf8}
\newcommand{\zh}[1]{\begin{CJK}{UTF8}{gbsn}#1\end{CJK}}

\usepackage{booktabs}

\usepackage{linguex}
\usepackage{enumitem}
\usepackage[htt]{hyphenat}
\usepackage{array}
\usepackage{xurl}
\usepackage{amsmath}
\usepackage{float}
\usepackage{multirow}
\usepackage{arydshln}

\title{A Cross-lingual Comparison of Human and Classification Model Entrainment Behavior in Code-switched Speech Settings}

\author{
  \textbf{Debasmita Bhattacharya\textsuperscript{1}}\thanks{Corresponding author: debasmita.b@cs.columbia.edu},
  \textbf{Siying Ding\textsuperscript{1}},
  \textbf{Alayna Nguyen\textsuperscript{2}}\thanks{Work done as an undergraduate at Columbia University.},
  \textbf{Julia Hirschberg\textsuperscript{1}}
\\
  \textsuperscript{1}Department of Computer Science, Columbia University \quad
  \textsuperscript{2}Instagram
}

\begin{document}
\maketitle
\begin{abstract}

Conversational entrainment is well-studied in monolingual and written contexts, but remains underexplored in spoken code-switching (CSW). We present a cross-lingual analysis of entrainment in Mandarin-English, Hindi-English, and Spanish-English dialogue, replicating and extending an established Spanish-English framework \cite{bhattacharya-etal-2024-measuring} to two new language pairs, and show that, while lexical entrainment generalizes across language pairs, entrainment over acoustic-prosodic and CSW style aspects exhibits context-specific variation. We build on these findings by asking whether classification models capture these human behavioral patterns. Using feature importance and ablation analyses, we find that classical and Transformer-based classifiers detect entrainment reasonably well but consistently prioritize features other than those most salient to human entraining behavior. Our approach introduces a human-grounded framework for evaluating model decision-making in multilingual stylistic contexts, and suggests challenges for developing  conversational agents capable of producing naturalistic code-switched speech.


\end{abstract}

\section{Introduction}

When speakers \textit{entrain} to one another, each subconsciously adapts aspects of the interlocutor's communication style into their own language production, coordinating features of diction, syntax, speaking rate, voice quality, pauses, and even facial expression and gesture, among others
\cite{Danescu_Niculescu_Mizil_2011, giles_coupland_coupland_1991, levitan11_interspeech, maurer-tindall, burgoon, Chartrand1999TheCE}. This interactive, multimodal phenomenon shapes conversational dynamics, enabling successful communication by ensuring mutual comprehension, fostering rapport, and making a positive impression on the interlocutor \cite{reitter-moore-2007-predicting, nenkova-etal-2008-high, levitan-etal-2012-acoustic}. 

While most existing work on entrainment has focused on monolingual, and especially English, communication, the last decade 
has increasingly seen entrainment studied in additional languages \cite{xia14_speechprosody, levitan-etal-2015-acoustic, kejriwal23b_interspeech, weise-etal-2021-talk} and in multilingual settings \cite{KOOTSTRA2010210, bawa-etal-2018-accommodation, parekh-etal-2020-understanding}.  
Among studies of \textit{code-switched entrainment}, where speakers adapt while alternating between producing at least two different language varieties, most analyses have focused on 
only a few language pairs, most often studying entraining behavior in Spanish-English \cite{soto18_interspeech, ahn-etal-2020-code, bhattacharya-etal-2024-measuring}. Though some studies have incorporated additional linguistic contexts and performed cross-lingual comparisons of lexical entrainment in written code-switching, few have done so in the spoken domain, giving rise to the present work. 

We explore the conversational dynamics of spoken code-switched dialogue in Mandarin-English, Hindi-English, and Spanish-English -- three CSW pairs that each involve English -- performing cross-lingual comparisons that validate and extend prior work, e.g., \citet{bhattacharya-etal-2024-measuring}, by grounding it in broader linguistic context. We first ask whether code-switched entrainment patterns in speech are consistent across language pairs involving different language families, levels of typological distance, and cultural settings. 
We then build upon this by asking whether entrainment detection models attend to the same features of entrainment as people do during spontaneous code-switched communication.
We find \textbf{1)} generally consistent multimodal patterns of entrainment across diverse language pairs, with slight variation reflecting linguistic and cultural idiosyncrasies of the languages involved; and \textbf{2)} while classifier models detect entrainment reasonably well across levels of conversational granularity and language pairs, they seem not to use the most relevant entraining features to do so, relative to humans. Our approach and findings suggest that empirically grounded human behavioral patterns provide a valuable lens for interrogating how and why classification models detect stylistic phenomena,
which has implications for understanding stylistic tasks beyond entrainment.

Our work is the first to cross-lingually validate entrainment patterns of spoken code-switching (CSW), and to use those patterns as a human-grounded framework for evaluating classifier behavior. In sum, our contributions are as follows: 

\begin{itemize}
    \item We replicate the Spanish-English entrainment approach applied in \citet{bhattacharya-etal-2024-measuring} across new Mandarin-English and Hindi-English settings, and extend it in a cross-lingual comparison of statistical entrainment patterns across three corpora. 
    \item We analyze models' feature importance attribution and ablation behavior to go beyond evaluating \textit{whether} models can detect entrainment, and also highlight \textit{why} they do so. 
    \item We release a manually annotated version of the MaSaC corpus with new CSW style labels to support further research on spoken CSW.\footnote{Data available at \url{https://tinyurl.com/2245wxe4}} 
\end{itemize}

\section{Related Work and Research Questions}

\paragraph{Entrainment in code-switched contexts.}Our work builds directly on \citet{bhattacharya-etal-2024-measuring}, who studied lexical, acoustic-prosodic, and multilingual stylistic entrainment in spontaneous Spanish-English conversations from the Bangor Miami (BM) corpus \cite{miami}. The authors applied the analytical framework of \citet{levitan11_interspeech}, which distinguishes entrainment at the turn-level from that at the conversation-level, and defines \textit{proximity}, \textit{convergence}, and \textit{synchrony} as unique manifestations of spoken entraining behavior (see Appendix~\ref{sec:app_method} for definitions). \citeauthor{bhattacharya-etal-2024-measuring} adapted this framework to study entrainment across CSW aspects and acoustic-prosodic features, while applying methods that draw on salient word classes and perplexity-based measures \cite{nenkova-etal-2008-high, gravano_tobi} to examine lexical entrainment in BM speech. They found that entrainment trends from monolingual domains appear in code-switched language production, with new patterns specific to spoken CSW, but did not examine whether these findings generalize beyond Spanish-English. Other work, such as \citet{bawa-etal-2018-accommodation} and \citet{parekh-etal-2020-understanding}, examined entrainment in Hindi-English CSW settings, focusing on lexical representations of speech transcripts and human-machine generated written exchanges, respectively. While both studies offered comparisons to entrainment patterns in Spanish-English CSW, these remained restricted to the written modality, giving rise to our first research question \textbf{RQ1:} \textit{To what extent do entrainment patterns of spoken CSW generalize across distinct language pairs?} To answer \textbf{RQ1}, we build on existing 
work on Spanish-English in BM, and additionally focus on Mandarin-English and Hindi-English; these four languages together represent the most spoken world languages \cite{ethnologue}. 

\paragraph{Comparing human and model multilingual behavior.}Recent work has proposed frameworks for explaining multilingual language models' ability to classify (and produce) human-like linguistic style 
and interaction dynamics \cite{namazifard-poech-2025-isolating, blevins-etal-2026-language, lin-etal-2025-v-vae}. We highlight 
\citet{havaldar-etal-2023-comparing}, who created multilingual politeness lexica and compared feature importance values derived from model encodings 
to human productions. Such work informs our second research question \textbf{RQ2:} \textit{To what extent do classifier models attend to those entraining features used by humans who perform entrainment in code-switched speech?} 
To answer \textbf{RQ2}, we examine a suite of classical and Transformer-based machine learning classifiers, using binary entrainment detection across language pairs to analyze feature importance in each 
setting.

\section{Corpora}

\begin{table*}[th]
\footnotesize
\centering
  \begin{tabular}{ ccccccc }
    \toprule
    \textbf{Corpus} &\textbf{Speech hours}  & \textbf{\# of dialogues} & \textbf{\# of utterances} & \textbf{\# of unique speakers} & \textbf{\% code-switched}\\
    \midrule
    Bangor Miami & 35 & 56 & 46,901 & 84 & 5 \\
    SEAME & 192 & 256
    & 110,146 & 156 & 52\\
    MaSaC & 13 & 1,190 & 6,476 &  5 & 54\\
    \bottomrule
  \end{tabular}
  \caption{Summary of relevant corpus statistics for each data set.}
    \label{tab:corpora}
    \vspace{-2mm}
\end{table*}

We examine three corpora of informal multilingual speech. Section~\ref{sec:part1_results} focuses on the \textbf{SEAME} corpus of spontaneous Mandarin-English dialogues \cite{seame} and the \textbf{MaSaC} corpus of Hindi-English situational comedy television speech \cite{masac}, comparing both to the \textbf{Bangor Miami} corpus of Spanish-English spontaneous conversations \cite{miami}.\footnote{See Appendix~\ref{sec:app_datalicenses} for data licensing details.} Section~\ref{sec:part2_results} analyzes all three corpora, which comprise recorded and transcribed monolingual and code-switched utterances from conversations between two or more speakers. 

The corpora cover shared topics, e.g., daily routines, 
interpersonal relationships, and academic/professional concerns. Unlike SEAME and BM, MaSaC does not consist of naturally-occurring speech. However, scripted television conversation is designed to sound natural and is therefore a suitable proxy for studying naturalistic phenomena like entrainment \cite{bawa-etal-2018-accommodation, danescu-niculescu-mizil-lee-2011-chameleons}, especially given the lack of other genre-matched, openly accessible speech data on the Hindi-English language pair.

Both BM and SEAME contain utterance-level annotations \cite{bhattacharya-etal-2024-measuring, bhattacharya-etal-2025-code} that identify code-switched and monolingual utterances, and distinguish between simple \textit{insertional} code-switches of single words or short phrases (\textit{I}), complex \textit{alternational} code-switches at grammatical clause boundaries (\textit{A}), and ``other'' (\textit{O}) strategies of CSW resembling tag-switching \cite{muysken-2000, poplack-1980}. We add the same sets of labels to MaSaC. First, we use Microsoft's open-source language identification (LID) tool\footnote{\url{https://github.com/microsoft/LID-tool}} to obtain token-level LID tags for the Romanized Hindi and English in MaSaC's speech transcripts, and accordingly infer utterance-level language labels: code-switched vs.\ monolingual English or Hindi. We then add manual annotations of CSW strategies on the multilingual utterances. All strategy labels are created by the first author and verified by a volunteer MS student, with no disagreements. Both have first-language proficiency in English; the former has intermediate competency in Hindi comprehension, while the latter is a native speaker of Hindi. We find that 92\% of code-switched MaSaC utterances use insertional CSW, 14\% use alternational, and less than 1\% use ``other'' CSW. This strategy distribution is comparable to those in BM (72\% \textit{I}, 13\% \textit{A}, 18\% \textit{O}) and SEAME (89\% \textit{I}, 12\% \textit{A}, 7\% \textit{O}) \cite{bhattacharya-etal-2024-measuring, bhattacharya-etal-2025-code}. We summarize additional corpus statistics in Table~\ref{tab:corpora}.

To enable direct comparison to known \textit{dyadic} entrainment patterns in 39 of the BM conversations \cite{bhattacharya-etal-2024-measuring}, we filter SEAME and MaSaC for conversations between just two speakers where speech audio and corresponding transcripts are available for both speakers. We retain only those containing 4 or more turns to ensure sufficiently long dialogue for conversation-level entrainment calculations (Section~\ref{sec:method_stats}). We accordingly use 40\% of SEAME (42,893 utterances over 33 dialogues) and 25\% of MaSaC (1,623 utterances over 334 dialogues) in our subsequent analyses.

\section{Methods}

\subsection{Computing code-switched entrainment.}
\label{sec:method_stats}

For statistical analyses of entrainment in Mandarin-English and Hindi-English CSW, we replicate \citet{bhattacharya-etal-2024-measuring}'s method, which we summarize here and detail fully in Appendix~\ref{sec:app_method}.



The \textbf{lexical} feature set comprises four word classes known to be often used and entrained upon in spontaneous dialogue \cite{nenkova-etal-2008-high, bhattacharya-etal-2024-measuring}:  \textit{corpus-level most frequent words} (top-100 and top-25), \textit{conversation-level most frequent words} (top-25), \textit{affirmative cues}, and \textit{fillers} (Appendix~\ref{sec:app_wordclasses} lists affirmative cues and fillers in each corpus). It also operationalizes broad similarity between speakers' \textit{overall language use} via perplexity from Kneser-Essen-Ney smoothed trigram language models trained on each speaker's transcripts with the KenLM toolkit. 
The \textbf{acoustic-prosodic} feature set covers utterance-level \textit{pitch}, \textit{intensity}, \textit{jitter}, \textit{shimmer}, \textit{harmonics-to-noise ratio} (HNR), and \textit{speaking rate} in syllables per second. All acoustic-prosodic features are extracted via Parselmouth \cite{parselmouth} with default parameter values and are $z$-score normalized by speaker. 
Utterance-level \textbf{CSW style} metrics include a binary flag for \textit{CSW presence}, a continuous \textit{CSW ratio} normalizing switching quantity by utterance length \cite{soto18_interspeech}, and a categorical label of the \textit{CSW strategy} in use (\textit{I}, \textit{A}, or \textit{O}), if any. 

We study entrainment over \textbf{acoustic-prosodic} and \textbf{CSW style} features via turn- and conversation-level proximity, convergence, and synchrony -- respectively, absolute similarity, change in similarity over time, and interlocutor coordination --  following the operationalizations of \citet{levitan11_interspeech} as adapted by \citet{bhattacharya-etal-2024-measuring}. We measure entrainment over \textbf{lexical} word classes using proximity and the stated 
perplexity approach for overall language use. 

\subsection{Evaluating model entrainment behavior.}
\label{sec:method_models}

We next examine how a suite of binary classifiers attends to turn- and conversation-level entrainment in spoken CSW with the goal of verifying whether, during classification, models assign the most importance to the conversational features over which humans significantly entrain. We use both classical machine learning methods and larger Transformer-based architectures.
The former group comprises eight classical models including \textit{SVC}, \textit{Decision Tree}, \textit{Random Forest}, and \textit{MLP} classifiers, all implemented in {\fontfamily{qcr}\selectfont
scikit-learn 1.6.1} (full list in Table~\ref{tab:hyperparams}). 
The latter group includes a 
baseline \textit{XGBoost} model, and adapted \textit{TabTransformer} \cite{huang2020tabtransformertabulardatamodeling}, \textit{FT-Transformer} \cite{ft-transformer}, \textit{XLM-R} \cite{xlm-roberta}, \textit{mBERT} \cite{devlin2019bertpretrainingdeepbidirectional}, \textit{RemBERT} \cite{chung2020rethinkingembeddingcouplingpretrained}, \textit{WavLM} \cite{Chen2021WavLMLS}, and \textit{mHuBERT} \cite{zanonboito24_interspeech} pre-trained contextual embedding models; we fuse each with an MLP for classification\footnote{We train only the MLP within each pre-trained fusion model, freezing the parameters of the pre-trained foundation.} 
and list their licenses in Appendix~\ref{sec:app_modellicenses}. 

To evaluate these models, we create a 75/15/10 train/validation/test stratified data split per corpus. 
For each classifier, we first perform grid search with 5-fold cross-validation, selecting the best hyperparameters by accuracy.\footnote{We list all hyperparameter details in Appendix~\ref{sec:app_hyperparams}.} Applying those in training, we
calculate the model's test-time feature importance values using
built-in attributes when available, and 
Shapley values \cite{SHAP} otherwise. 
We then perform feature-level ablations with the same (best) model and 5-fold cross-validation to indirectly evaluate the importance of each feature of entrainment from Section~\ref{sec:method_stats}.

At the \textbf{conversation-level}, the positive (entraining) class comprises  conversations exhibiting significant proximity on an empirically determined feature set of interest:\footnote{Based on Section~\ref{sec:part1_results}, only the feature set of interest should discriminate entraining from non-entraining speech: other feature sets show entrainment in either nearly all or nearly no instances, providing little discriminative signal. This data-dependence is a property of the empirical entrainment patterns reported in Section~\ref{sec:part1_results} rather than a fixed methodological choice made in advance, and we adopt the same feature-set determination procedure uniformly across corpora. \label{footnote:5}} all CSW style metrics (BM, SEAME) or 
lexical word classes (MaSaC); the negative class is its complement. 
Classes per corpus are roughly balanced. At the \textbf{turn-level}, the positive class is similarly empirically determined (Section~\ref{sec:part1_results}) and comprises consecutive turn pairs with significant proximity on a majority of CSW style metrics (BM, SEAME) or acoustic-prosodic features (MaSaC), with class imbalance handled via oversampling.
At both levels of conversational granularity, a model input is a feature vector covering all feature sets from Section~\ref{sec:method_stats} for each speaker, with continuous features averaged and non-continuous features first converted to binary flags; full construction details are in Appendix~\ref{sec:app_posnegclasses}. This results in a \textit{tabular}/\textit{numerical} representation, which we concatenate with model-specific embeddings of conversation transcripts (text embeddings) or audio recordings (speech embeddings). 

\section{Results}

\subsection{Cross-lingual human entraining behavior.}\label{sec:part1_results}

We first examine whether human entrainment patterns over lexical, acoustic-prosodic, and CSW style features in code-switched speech are consistent across language pairs differing in typological distance, language family, and cultural context. Table~\ref{tab:illustrative-examples} shows representative examples in each case. We report effect sizes and Bonferroni-corrected significance results 
in Appendices~\ref{sec:app_effect_sizes} and \ref{sec:app_bonferroni}.

\paragraph{Entrainment on lexical aspects of code-switched speech.}Within SEAME, we find strong lexical entrainment on all word classes. Over the most frequent corpus-level 
words, 100\% of SEAME conversations exhibit entrainment ($p < 0.05$ in each case). We find similar entrainment on affirmative cues (97\% of conversations, $p < 0.05$) and fillers (91\% of conversations, $p < 0.05$). In terms of overall language use, when including out-of-vocabulary (OOV) words, 88\% of conversations exhibit within-conversation entrainment ($p < 0.05$); 77\% involve both interlocutors entraining to their partner's overall language use, while in the remaining 11\% only one speaker entrains to the other. Results excluding OOV words are nearly identical (Appendix~\ref{sec:app_acoustic-prosodic-seame}).

Within MaSaC, we find significant lexical entrainment on the word classes involving frequent words. Over the corpus-level most frequent words (top-100 and top-25), 84\% and 92\% of MaSaC conversations respectively exhibit entrainment ($p < 0.05$ for both). While entrainment over the top-25 conversation-level words is significant ($p < 0.05$), it appears only in 10\% of conversations. In terms of overall language use, 70\% of conversations show evidence of within-conversation entrainment when including OOV words ($p < 0.05$): 52\% exhibit bidirectional entrainment while 18\% involve only one speaker entraining to the other. OOV-excluded results are almost identical (Appendix~\ref{sec:app_acoustic-prosodic-seame}). 

\begin{figure}
\includegraphics[width=0.9\columnwidth]{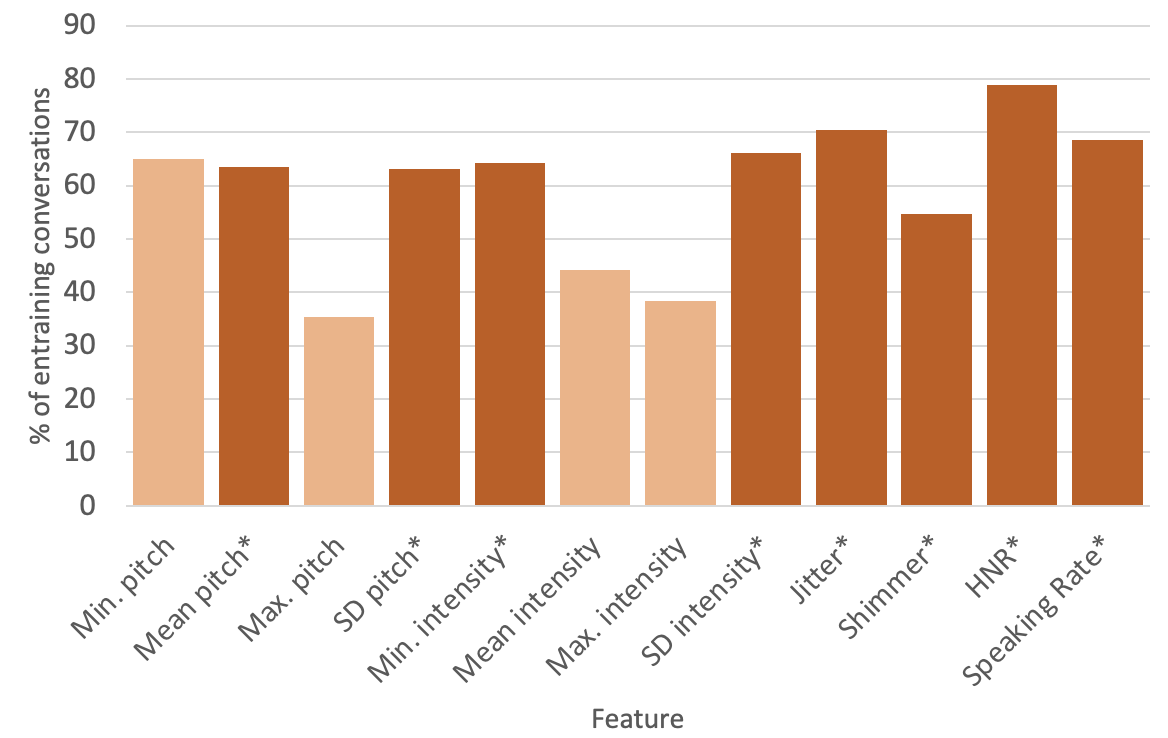}
    \caption{Turn-level acoustic-prosodic proximity in MaSaC. * and darker bars indicate significant features.}  
    \label{fig:proximity-turn-masac}
    \vspace{-2mm}
\end{figure}
\begin{figure}
\includegraphics[width=0.9\columnwidth]{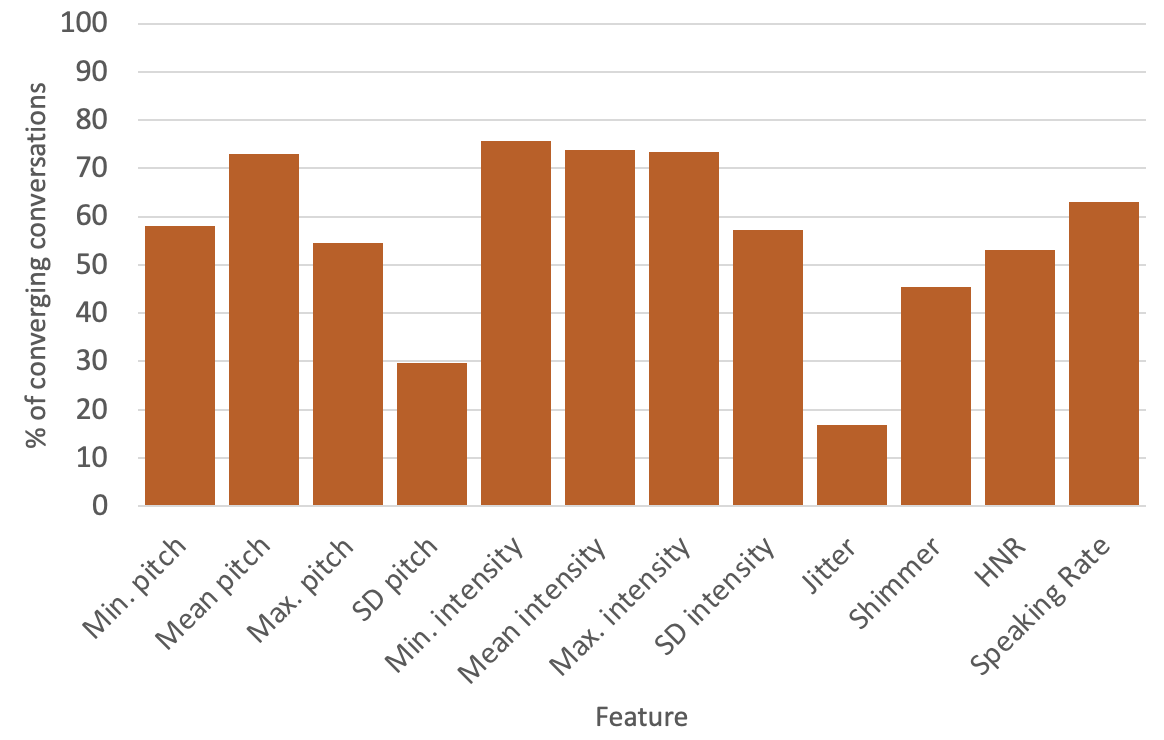}
    \caption{Turn-level acoustic-prosodic convergence in MaSaC. Most conversations converge \textit{strongly}; non-converging ones diverge \textit{weakly}. All features significant.}  
    \label{fig:convergence-turn-masac}
    \vspace{-3mm}
\end{figure}

\paragraph{Entrainment on acoustic-prosodic aspects of code-switched speech.}We find limited acoustic-prosodic entrainment within SEAME. Most conversations exhibit turn-level proximity and synchrony, but these achieve statistical significance on fewer than half of the acoustic-prosodic features (Figures~\ref{fig:proximity-turn-seame} and \ref{fig:synchrony-turn-seame} in Appendix~\ref{sec:app_acoustic-prosodic-seame}), with uniformly small effect sizes across tested turn-level features ($|d| < 0.04$; Appendix~\ref{sec:app_effect_sizes}). Similarly, significant turn-level convergence occurs in half of SEAME conversations over only a few pitch and intensity features (Figure~\ref{fig:convergence-turn-seame} in Appendix~\ref{sec:app_acoustic-prosodic-seame}), suggesting inconsistent acoustic-prosodic entrainment of this kind.  
In MaSaC, in contrast, we find much stronger evidence of acoustic-prosodic entrainment. In terms of turn-level proximity, more than 60\% of conversations entrain significantly over two-thirds of acoustic-prosodic features (Figure~\ref{fig:proximity-turn-masac}). The majority of conversations also consistently converges strongly across features at the turn-level (Figure~\ref{fig:convergence-turn-masac}). 

\paragraph{Entrainment on CSW style aspects of code-switched speech.}
We find significant proximity at both the turn-level (67\% of conversations, $p < 0.05$) and the conversation-level (97\% of conversations, $p < 0.05$) for CSW presence in SEAME, as well as weak, but 
statistically insignificant, turn-level synchrony in 67\% of conversations. In terms of CSW ratio, we again find significant turn- and conversation-level proximity in 70\% and 97\% of conversations, respectively ($p < 0.05$ in both). As with CSW presence, there is weak turn-level synchrony over CSW ratio in 91\% of conversations, though this, too, fails to reach statistical significance. Finally, examining CSW strategies, we find significant turn-level synchrony over \textit{insertional} CSW (67\% of conversations, $p< 0.05$), and significant conversation-level proximity over \textit{alternational} CSW (88\% of conversations, $p < 0.05$). Effect sizes reinforce granular asymmetry: SEAME's turn-level CSW style proximity effects are consistently weak, while the corresponding conversation-level effects are medium-to-large (Appendix~\ref{sec:app_effect_sizes}).


Our findings on MaSaC are strikingly different from those on SEAME. We find only significant turn-level convergence across metrics of CSW style: 57\% of conversations converge on CSW presence (54\% \textit{strongly}, 3\% weakly or moderately), 91\% converge on CSW ratio (82\% \textit{strongly}, 9\% weakly or moderately), and 62\% converge (\textit{strongly}) on \textit{insertional} CSW. Across metrics, this convergence is consistently statistically significant. Since MaSaC's scripted nature limits our ability to disentangle whether these patterns reflect Hindi-English-specific entrainment norms or corpus- and genre-specific effects (see Discussion below), we scope these findings to the MaSaC corpus only.

\begin{table*}[t]
\centering
\footnotesize
\setlength{\tabcolsep}{3pt}
\renewcommand{\arraystretch}{0.85}
\begin{tabular}{@{}p{1.1cm}p{2.1cm}p{8cm}p{2.1cm}p{1.9cm}@{}}
\toprule
\textbf{Corpus} & \textbf{Type} & \textbf{Exchange} & \textbf{Feature} & \textbf{Value(s)} \\
\midrule
SEAME & Lexical & \textbf{S1:} \textit{\zh{对} then \zh{我 就} like yah \zh{我 那 时候}} [\textit{Right, then I was like, yeah, at that point I...}] \textbf{/} \textbf{S2:} \textit{then \zh{你 就 过 不 了}} [\textit{...then you wouldn't be able to get across.}] & Frequent word  & `\textit{then}' \\
MaSaC & Lexical & \textbf{S1:} \textit{Chalo, chalo, come on, Paris nahi jaana!} [Come on, come on, come on, I'm not going to Paris!] \textbf{/ S2:} \textit{Nahi, nahi, nahi!} [\textit{No, no, no!}] & Frequent word & `\textit{nahi}' \\
SEAME & Acoustic-prosodic, non-entraining\ & \textbf{S1:} \textit{what what what \zh{什么}?} [\textit{What what what what?}] \textbf{/ S2:} \textit{do you think our hall has pretty girls?} & SD pitch (Hz) & 211.5 vs.\ 12.4 \\
MaSaC & Acoustic-prosodic, entraining\ & \textbf{S1:} \textit{That's totally ridiculous Indravardhan} \textbf{/ S2:} \textit{Arey Maya...yakin karo} [\textit{Oh Maya...believe me}] & Mean pitch (Hz) & 231.3 vs.\ 231.4 \\
SEAME & CSW style\ & \textbf{S1:} \textit{\zh{对 那个} area five \zh{多少}} [\textit{right, that Area Five, how much is it?}] \textbf{/ S2:} \textit{okay kay \zh{让我拿出我的} calculator} [\textit{Okay, okay, let me take out my calculator}] & CSW presence & --- \\
MaSaC & CSW style\ & \textbf{S1:} \textit{Poore nau rupay waste!} [\textit{All 9 rupees wasted!}] \textbf{/ S2:} \textit{...Pachhatar paise vaala pen hai ye?} [\textit{...Is this a 75-paise pen?}] & CSW strategy: \textit{I} & --- \\
\bottomrule
\end{tabular}
\caption{Turn-level examples of entrainment (and disentrainment) as discussed throughout Section~\ref{sec:part1_results}.}
\label{tab:illustrative-examples}
\vspace{-2mm}
\end{table*}



\paragraph{Discussion.}
\citet{bhattacharya-etal-2024-measuring} previously found that monolingual entrainment trends extend to Spanish-English CSW in BM, alongside new patterns specific to CSW itself (e.g., entrainment on CSW presence, ratio, and strategy), but did not test whether these findings generalize beyond Spanish-English. Compared to this prior work, we generally find consistent \textit{lexical} patterns across language pairs, robust to Bonferroni correction both across and within corpora (Appendix~\ref{sec:app_bonferroni}).
Both Mandarin-English and Spanish-English CSW exhibit lexical entrainment over all word classes. In Hindi-English, entrainment is absent over conversation-level frequent words, affirmative cues, and fillers. We attribute the former absence to shorter 
conversations in MaSaC limiting its scope: fewer dialogue turns and shorter utterances (Table~\ref{tab:corpora}) leave less opportunity for a stable high-frequency word set to emerge and for speakers to entrain on them.  
We suspect the latter two null results are due to these lexical categories not appearing in scripted MaSaC dialogue, an entrainment effect noted in other scripted speech settings \cite{blevins-etal-2026-language}; their effect sizes are near zero (Appendix~\ref{sec:app_effect_sizes}), consistent with a genuine absence of entrainment rather than low statistical power. This underscores a general caveat: MaSaC findings may reflect corpus properties as much as linguistic ones, and replication on naturally occurring Hindi-English speech is necessary before strong conclusions can be drawn. Entrainment on overall language use, in contrast, is robust across all three corpora (SEAME, MaSaC, and BM).

In terms of \textit{acoustic-prosodic} entrainment, both Mandarin-English and Hindi-English mirror known Spanish-English patterns on the \textit{absence} of conversation-level proximity and convergence, and the \textit{absence} of significant turn-level synchrony. Where entrainment is \textit{present}, MaSaC's Hindi-English patterns align more closely with BM's Spanish-English ones than do those from SEAME's Mandarin-English, which lacks acoustic-prosodic entrainment. One explanation is Mandarin's unique tonal character and distinct intonation norms relative to Spanish and Hindi, though corpus-level differences in speaker population and interaction setting may also contribute slightly.


With respect to \textit{CSW style}, the Mandarin-English patterns we find align strongly with those known for Spanish-English: 
speech in both language pairs exhibits entrainment over most multilingual features of CSW conversation via proximity and synchrony. MaSaC's Hindi-English CSW differs markedly, exhibiting only turn-level convergence, and less prominent entrainment over CSW style metrics overall. 
We offer a preliminary hypothesis for this difference. Hindi-English speakers exhibit strikingly higher M- and I-index \cite{m-index,i-index} values ($M=0.49$; $I=0.48$) than Mandarin-English ($M=0.14$; $I=0.13$) or Spanish-English ($M=0.08$, $I=0.01$) speakers, indicating more abundant and frequently interleaved CSW. One interpretation is that when speakers already code-switch prevalently and independently, there is less room for one speaker's style to influence the other's, leading to a ceiling on stylistic entrainment. We do not independently test this hypothesis and note that  
disentangling possible contributions of CSW linguistic norms and speech genre would require comparing naturally-occurring Hindi-English dyads across several CSW rates, which we leave for future work.


In response to \textbf{RQ1}, \textit{lexical} entrainment in spoken CSW generalizes robustly across language pairs, while \textit{acoustic-prosodic} and \textit{CSW style} patterns exhibit variation that may reflect the unique linguistic character of the non-English language and its associated cultural norms, as well as potential corpus-specific interactional factors.  We offer further qualitative comparison to prior monolingual and CSW entrainment findings in Appendix~\ref{sec:app_corpus_comparisons}.



\subsection{Classifier models' entrainment behavior.}\label{sec:part2_results}

We next investigate how classifiers behave when detecting entrainment, relative to human entraining behavior from multiple CSW contexts and levels of conversational granularity. We apply both model-internal and model-external evaluation methods, first inspecting alignment between the most important features to the best-performing models\footnote{For conversation-level feature importance, we examine only models whose performance exceeds a test-time accuracy of 75\% on BM and SEAME, and 65\% on MaSaC; for the turn-level we use a 65\% threshold across corpora. These maintain the reliability of feature importance observations. Slightly lower thresholds for MaSaC and turn-level analyses account for substantially smaller or noisier datasets; higher thresholds would overly restrict qualitative feature interpretation.} and those known to be statistically salient for entrainment per corpus, and then examining the impact of feature-level ablations across all models. Table~\ref{tab:classifier_accuracy} summarizes best-performing classifier accuracy by corpus and conversational granularity; additional per-model details are in Figures~\ref{fig:bm-feature-importances-full}-\ref{fig:masac-feature-importances-turnlevel}. We visualize conversation-level results for BM; SEAME and MaSaC figures are in Appendices~\ref{sec:app_feature_importances} and~\ref{sec:app_ablations}.

Since each corpus empirically differs in which feature set discriminates entraining from non-entraining speech (Footnote~\ref{footnote:5}), the specific classification task differs by corpus; 
cross-corpus comparisons of model behavior should be interpreted with this 
in mind. Note that statistical salience in human entrainment analyses need not reflect causal importance in such conversations, and classifiers' feature importance estimates may be influenced by the task operationalization. We examine alignment between model predictive behavior and our statistical characterization of human entrainment, 
rather than underlying cognitive mechanisms.

\begin{table}[t]
\centering
\small
\begin{tabular}{@{}llp{5cm}@{}}
\toprule
\textbf{Corpus} & \textbf{Level} & \textbf{Classifiers (Accuracy)} \\
\midrule
\multirow{2}{*}{BM} & Conv. & \textit{Decision Tree}$^{*}$ (0.99), \textit{Random Forest}$^{*}$ (0.99), \textit{XGBoost}$^{\dagger}$ (0.99) \\ \cdashline{2-3}
\noalign{\vskip 3pt}
 & Turn & \multicolumn{1}{l}{Excluded (see Footnote~\ref{footnote:10})} \\
\midrule
\multirow{2}{*}{SEAME} & Conv. & \textit{RemBERT+WavLM fusion}$^{\dagger}$ (0.99), \textit{MLPClassifier}$^{*}$ (0.99), \textit{XLM-fusion}$^{\dagger}$ (0.99), \textit{mBERT-fusion}$^{\dagger}$ (0.99), \textit{RemBERT-fusion}$^{\dagger}$ (0.99), \textit{GradientBoostingClassifier}$^{*}$ (0.99), \textit{GaussianNB}$^{*}$ (0.99), \textit{TabTransformer-adapted}$^{\dagger}$ (0.99), \textit{FT-Transformer}$^{\dagger}$ (0.75) \\ \cdashline{2-3} \noalign{\vskip 3pt}
 & Turn & \textit{mBERT-fusion}$^{\dagger}$ (0.70), \textit{RemBERT-fusion}$^{\dagger}$ (0.67), \textit{TabTransformer-adapted}$^{\dagger}$ (0.65) \\
\midrule
\multirow{2}{*}{MaSaC} & Conv. & \textit{LogisticRegression}$^{*}$ (0.71), \textit{mBERT-fusion}$^{\dagger}$ (0.68), \textit{SVC}$^{*}$ (0.65) \\ \cdashline{2-3} \noalign{\vskip 3pt}
 & Turn & \textit{mBERT+WavLM-fusion}$^{\dagger}$ (0.77), \textit{WavLM-fusion}$^{\dagger}$ (0.74), \textit{mBERT-fusion}$^{\dagger}$ (0.70), \textit{mHuBERT-fusion}$^{\dagger}$ (0.70), \textit{XLM-fusion}$^{\dagger}$ (0.69) \\
\bottomrule
\end{tabular}
\caption{Best-performing $^{*}$classical and $^{\dagger}$Transformer-based classifiers by corpus \& conversational granularity.}
\label{tab:classifier_accuracy}
\vspace{-2mm}
\end{table}

\begin{figure}
\centering
\includegraphics[width=0.95\linewidth]{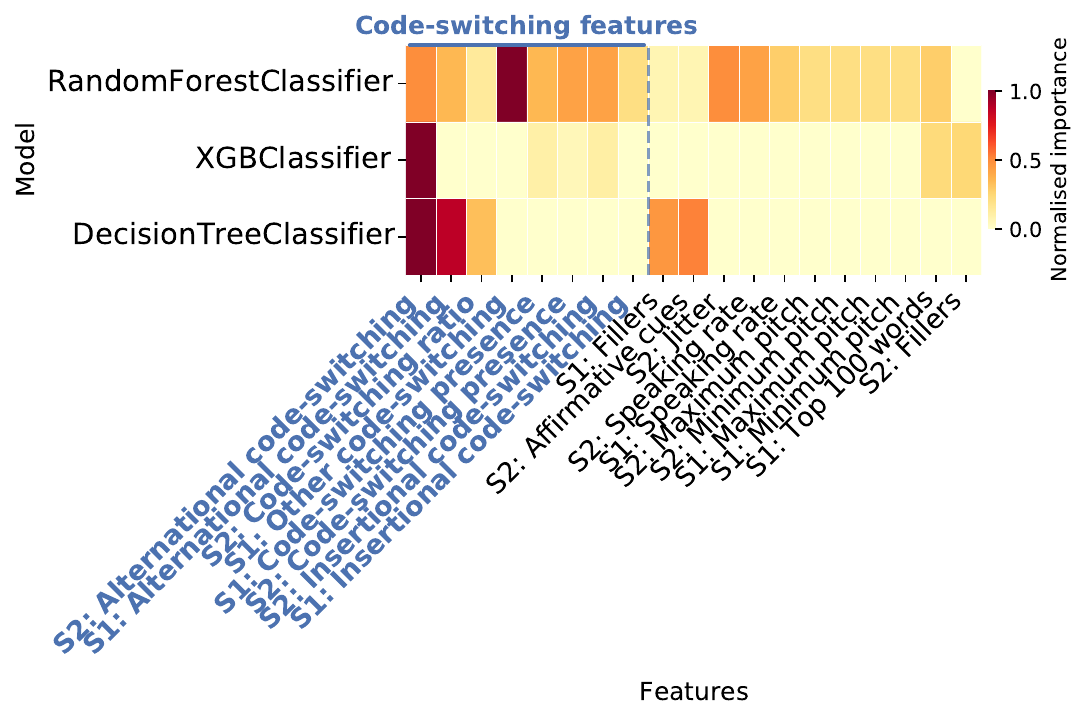}
    \caption{Highest feature importance values across best-performing BM models at the conversation-level. We normalize values per model using row-wise min.-max. so that models with different importance scales are comparable. For visualization across all features, including many near-zero values, see Figure~\ref{fig:bm-feature-importances-full} in Appendix~\ref{sec:app_feature_importances}.
    }  
    \label{fig:bm-feature-importances}
    \vspace{-3mm}
\end{figure}

\paragraph{Models' feature importance values.}At the \textbf{conversation level} of BM, 3 models 
achieve near perfect accuracy on distinguishing dialogue in which speakers entrain on CSW style features from that lacking such entrainment: \textit{Decision Tree},\footnote{criterion: gini, max. depth: 3, min. samples split: 2} \textit{Random Forest},\footnote{criterion: log loss, max. depth: 1, min. samples split: 5} and \textit{XGBoost}.\footnote{booster: gbtree, lr: 0.01, max. depth: 3, \# estimators:  50} Each assigns the greatest feature importance to at least one relevant CSW style metric (Figure~\ref{fig:bm-feature-importances}), yet two unexpected patterns emerge across these models. First, both \textit{DecisionTree} and \textit{XGBoost} concentrate importance on a \textit{single} CSW feature rather than across the full set of human-relevant ones; this is surprising given that positive-class conversations exhibit significant entrainment on \textit{each} CSW feature. We also note the assignment of relatively high importance to acoustic-prosodic features, particularly in the \textit{RandomForest}; these should provide little signal for distinguishing conversations with and without entrainment over CSW aspects, given that nearly all BM conversations exhibit significant conversation-level acoustic-prosodic entrainment \cite{bhattacharya-etal-2024-measuring}, and these do not correlate with CSW features (details in Table~\ref{tab:conv_level_corr} in Appendix~\ref{sec:app_correlations}).

\begin{figure*}
\centering
\includegraphics[width=0.9\linewidth]{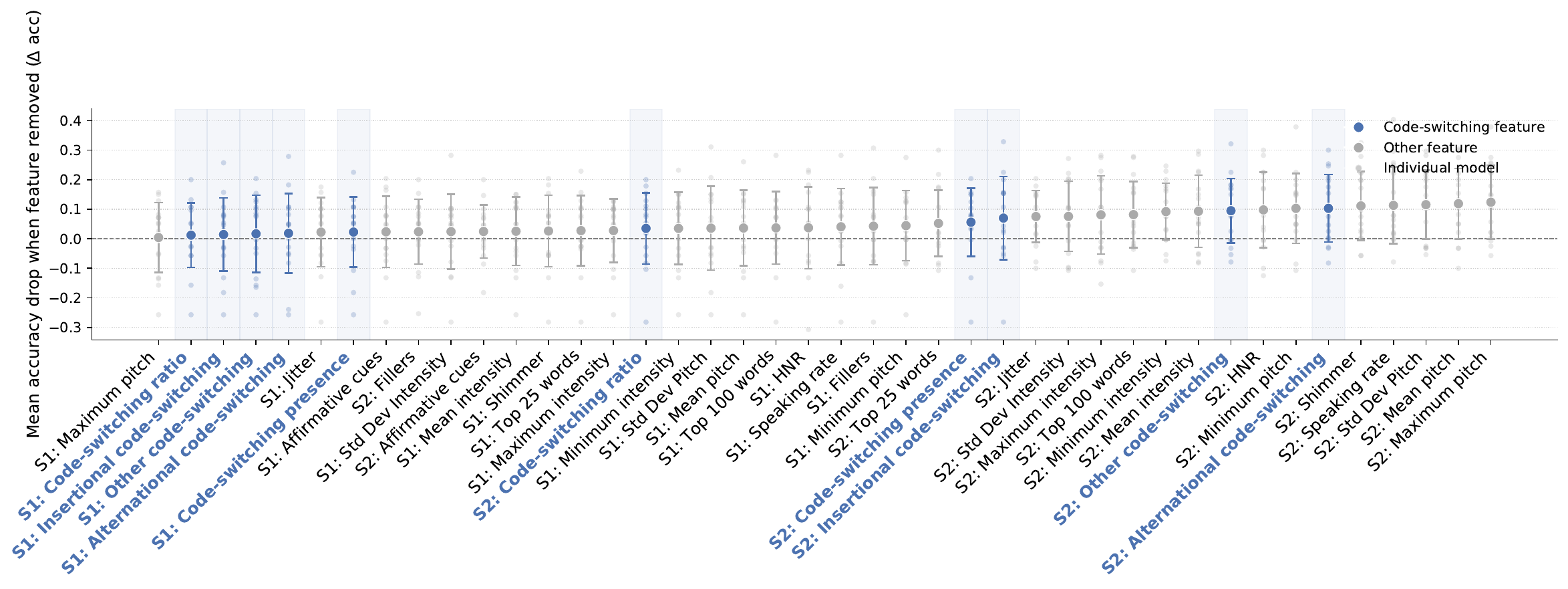}
    \caption{Aggregated feature ablation results for BM at the conversation-level. Features are ordered left to right by increasing impact; 
    $\Delta_{acc.}>0$ means accuracy \textit{drops} when the feature is removed, i.e., the feature is useful. 
    }  
    \label{fig:bm-ablations}
    \vspace{-2mm}
\end{figure*}

These alternative predictive patterns replicate across corpora.
All but one top-performing SEAME classifier assigns the greatest importance to features outside the relevant CSW style set (Figure~\ref{fig:seame-feature-importances}); the best MaSaC models do the same for the relevant lexical set (Figure~\ref{fig:masac-feature-importances}). Across corpora, we find little evidence that \textbf{conversation-level} classification decisions are guided by the features most expected from human entrainment behavior. 

At the \textbf{turn level}, several models surpass the classification accuracy threshold with respect to CSW style features on SEAME and acoustic-prosodic features on MaSaC (Table~\ref{tab:classifier_accuracy} and Appendix~\ref{sec:app_feature_importances}).\footnote{No BM model exceeds the 65\% threshold, so we exclude BM from turn-level analysis; discussion in Appendix~\ref{sec:app_BM_exclusion}. \label{footnote:10}} For SEAME, none assigns meaningful importance to any relevant CSW style features (Figure~\ref{fig:seame-feature-importances-turnlevel}), instead concentrating on acoustic-prosodic cues, which are not among those that SEAME speakers use to significantly entrain at the turn level (Section~\ref{sec:part1_results}), nor do these correlate with CSW features (Table~\ref{tab:turn_level_corr};  Appendix~\ref{sec:app_correlations}). This mirrors the conversation-level for SEAME, where the best-performing models similarly bypass relevant CSW style features in favor of acoustic-prosodic signal. 
For MaSaC, we observe a divergence by model type. Text-embedding models show the same concentration effect noted for conversation-level BM, assigning the greatest importance to a narrow subset of pitch and intensity cues while making limited use of the broader acoustic-prosodic profile characterizing human entraining behavior (Figure~\ref{fig:masac-feature-importances-turnlevel}). Speech-embedding models, in contrast, distribute importance across the full acoustic-prosodic feature set, but also assign notably high importance to CSW style features, which do not define the positive class for this task; turn pairs are labeled by acoustic-prosodic proximity, not CSW style entrainment. We propose an explanation for this cross-feature-set 
pattern
in Appendix~\ref{sec:app_feature_importances}.

These \textbf{turn-level} results largely reinforce those from the conversation level in their lack of sufficient evidence for model classification decisions being systematically guided by the features most salient to human entrainment. While speech-grounded representations offer partial improvement in feature coverage, these exhibit diffuse rather than targeted reliance on human-grounded turn-level entrainment.
We also note that model performance is generally lower at the turn level than at the conversation level across corpora, a pattern consistent with the finer-grained and more variable nature of turn-level conversational dynamics, which may reduce the separability of entraining and non-entraining instances. The relative advantage of contextual embedding-based models at the turn level 
is consistent with this interpretation: contextual representations may provide useful inductive bias when local sequential context is more diagnostic than aggregate conversational statistics.

\paragraph{Feature-level ablations.}Since feature importance estimates are
classifier-specific and derived from only a subset of models, we expand our analysis to all models via feature-level ablations. At the \textbf{conversation level} of BM, ablating each feature produces a significant \textit{drop} in accuracy relative to the baseline, per paired \textit{t}-tests; all features thus provide some useful signal (Figure~\ref{fig:bm-ablations}). However, most effect sizes are small, and no relevant CSW style features rank among the top contributors to accuracy drops. SEAME ablations are similar and 
 generally 
produce negligible change in performance (Figure~\ref{fig:seame-ablations}). While CSW style features account for 3 of the top-5 contributors to performance, more are linked to accuracy \textit{increases} upon ablation, suggesting that they introduce noise rather than useful signal. This effect is more pronounced in MaSaC: only one lexical feature ablation results in an accuracy drop, and none rank among the top contributors to 
performance across models (Figure~\ref{fig:masac-ablations}). 

We find a larger divergence between model and 
human entraining behavior at the \textbf{turn level}. For SEAME and BM, ablations produce near-zero changes in accuracy (Figures~\ref{fig:seame-ablations-turnlevel} and \ref{fig:bm-ablations-turnlevel}), suggesting that models rely on diffuse (or redundant) cues 
rather than any identifiable set of uniquely important entrainment features to make decisions. 
MaSaC models are only slightly more informative: while some salient acoustic-prosodic features contribute to performance, effect sizes remain small and importance is narrowly concentrated, with only speaking rate as a partial exception (Figure~\ref{fig:masac-ablations-turnlevel}). 
These results suggest that, as conversational granularity increases, entrainment becomes harder for models to capture in a way that reflects statistically observed human entrainment patterns.


\paragraph{Discussion.}Across language pairs and evaluation methods, we observe two recurring patterns in model behavior. At the conversation-level, models achieve reasonable performance while relying on features different from those identified as statistically salient in human entrainment. At the turn-level, feature ablations produce near-zero changes, suggesting reliance on diffuse 
signals rather than a coherent, human-grounded feature set. 
In response to \textbf{RQ2}, 
our entrainment detection models 
generally prioritize features other than those most salient
in naturalistic CSW, 
and, at finer-grained levels of interaction,  either distribute importance too broadly across feature sets or place limited emphasis on
any interpretable, human-relevant features. Similar to \citet{blevins-etal-2026-language}, 
we speculate that this may reflect fundamental differences between underlying human and model entrainment mechanisms, as well as the inherent difficulty of explicitly modeling a \textit{subconscious} human process. Other explanations include a need for alternative representations of entraining code-switched speech, or the possibility of our data and models capturing latent aspects of entrainment not covered by the formal theories used to operationalize our approach. Further investigation of these factors, particularly on how models might be encouraged to attend to and robustly represent human-grounded entrainment features, is an important direction for future work. 


As noted in prior work \cite{havaldar-etal-2023-comparing},  multilingual 
models generally classify stylistic aspects of language well, but struggle to generate them appropriately \cite{briakou-etal-2021-ola, srinivasan-choi-2022-tydip}. 
Our findings raise the possibility that generating code-switched speech with human-like entrainment patterns may pose a similar challenge, though results from classifier-based settings may not directly transfer to generative models. One potential concern is that classifiers, which show promise for steering and controlling generative models' stylistic outputs \cite{Dathathri2020Plug, scalena-etal-2026-steering, dinan-etal-2020-multi, liang-etal-2024-controlled}, could propagate feature-level misalignment when used to guide generation. So, strong performance in detecting entrainment need not imply that models capture the underlying phenomenon in a way that would support naturalistic generation.
 Recent findings in monolingual settings reinforce this concern: \citet{blevins-etal-2026-language} identified mismatches in the 
 linguistic form and motivations of entrainment between humans and language models in spontaneous text production, while \citet{Hua2025:post-train} showed that post-training fails to close the gap between LLMs and humans in task-oriented stylistic adaptation. Taken together, these results suggest that it may be valuable to directly investigate whether generative models 
 face similar challenges in the more demanding context of code-switched speech,
 and whether interventions (e.g., prompting or training strategies) can encourage more human-like entrainment behavior.

\section{Conclusion}

We explore 
cross-lingual entrainment behavior in spoken CSW, studying both human speakers and classification models across language pairs spanning diverse typological families and cultural contexts. We find that many established patterns of code-switched entrainment extend to new language pairs, while observed cross-corpus variation may reflect relevant multilingual norms. We also find that the predictive cues used by our entrainment classifiers are not well aligned with statistically salient features observed in human speech under our operationalization
-- a finding we surface through feature importance attribution methods that are directly tied to model behavior and enable human-interpretable explanations. We hope that these new insights, together with the stylistically annotated data we release, will inform the development of multilingual language models capable of generating naturalistic code-switched speech that embodies the human-like stylistic patterns needed for successful communication and social cohesion. 


\section*{Limitations}

Our work is the first to conduct a cross-lingual analysis of entrainment in code-switched speech. While we study multiple language pairs covering distinct language families and cultural contexts, all involve English as one of the languages. This design choice is partly motivated by data availability constraints, but necessarily leaves much of the world's rich and varied CSW landscape unaddressed. Our claims about cross-lingually consistent entrainment patterns are therefore limited to English-involved CSW, and may not extend to bilingual settings excluding English. 

We also acknowledge that our corpora each represent a specific community of CSW speakers (Miami-based Hispanic communities, Singaporean and Malaysian Mandarin-speakers, one particular Hindi-language television show), in addition to differing in size, genre, speaker demographics, and overall CSW rate (Table~\ref{tab:corpora}). Fully disentangling language-pair-specific entrainment patterns from the corpus-level differences we find would require new, genre- and size-matched corpora across all three language pairs, which do not currently exist; collecting or identifying such data, particularly a naturally-occurring Hindi-English counterpart to MaSaC, is a concrete direction we intend to pursue in future work. Since models are highly context-sensitive and entrainment norms may vary within a language pair across regions, social contexts, and speaker demographics, we eagerly anticipate future work that tests the robustness of our findings on new data. Given limited access to suitable open-source code-switched speech corpora, our work is a first step towards expanding cross-lingual understanding of conversational dynamics in CSW. 


Separately, we acknowledge a potential data pre-processing limitation specific to MaSaC. The Romanized representation of Hindi in its speech transcripts may have posed a challenge for the LID tool used to obtain token-level language tags, as language-ambiguous words can be difficult to label automatically without contextual or pronunciation information. We manually reviewed and corrected LID labels prior to inferring utterance-level language assignments, reading and listening to \textit{every} 
MaSaC transcript in full, but some residual errors could remain. 
If any, these would most likely result in underestimating the number of code-switched utterances, and their corresponding style annotations, by miscategorizing them as monolingual. This is a particularly acute concern for MaSaC given its limited size. This limitation reflects a broader challenge of computational work on Hindi-English CSW: the scarcity of openly accessible, genre-matched spoken corpora means that errors in automatic pre-processing carry greater consequence, as few alternative data sources exist. 

Finally, we acknowledge three related limitations in our model evaluation methodology. First, our binary classification framing
is necessarily a simplification of what is a continuous and multidimensional linguistic phenomenon. The thresholds used to construct entraining and non-entraining classes directly shape what models are able to learn, and different operationalizations of entrainment could yield different classification boundaries and, consequently, different feature importance profiles. Since our entrainment labels are derived from the same analytical framework used to characterize human behavior, the resulting classification task necessarily reflects that operationalization of entrainment rather than the full underlying phenomenon. 

Second, we operationalize features salient to human entraining behavior as those exhibiting statistically significant entrainment patterns (Section~\ref{sec:part1_results}). We note that statistical significance in corpus-level analysis is an imperfect proxy for behavioral or cognitive salience: a feature may show significant entrainment at the population level without being a primary driver of individual speakers' adaptation, and vice versa. Our comparison between human and model feature reliance should therefore be understood as a first approximation, i.e., a way of asking whether models are at least sensitive to the same dimensions of speech that correlate with entrainment, rather than a direct test of the underlying mechanism. Future work using behavioral data, e.g., listener judgments or physiological measures of rapport, could provide a more direct index of which features matter more to human speakers.

Third, Shapley feature importance estimates, which we rely on for a subset of models in RQ2, are known to be sensitive to feature correlations; since our acoustic-prosodic features are weakly to moderately correlated with one another (Tables~\ref{tab:conv_level_corr} and \ref{tab:turn_level_corr}), some instability in their attributed importances is possible. Both of these factors introduce uncertainty into the comparison between human and model entrainment behavior, and future work would benefit from exploring alternative operationalizations and importance attribution methods to assess the robustness of our conclusions.

\section*{Ethical Considerations}

This study uses only secondary data and does not involve human experiments. All speakers in the Bangor Miami and SEAME corpora consented to data collection and sharing, and both datasets were released in de-identified form by their original authors. The MaSaC corpus consists of fictional dialogue performed by professional actors and was used with permission from its creators.

While our work is primarily exploratory, it has potential implications for multilingual conversational systems. Modeling conversational entrainment in code-switched settings may support applications such as adaptive dialogue systems or language learning tools, but could also be misused in contexts where stylistic adaptation is intended to increase trust or rapport, including persuasive or deceptive scenarios such as targeted scams or disinformation. Care should therefore be taken in the deployment of downstream applications.

Our findings highlight how entrainment patterns vary across language pairs and contexts, and systems trained on such data may reflect these underlying norms. This may lead to uneven performance or misalignment when applied to speakers whose communication style differs from those represented in these datasets, and raises broader questions about whether entrainment should be treated as descriptive behavior or a normative benchmark.

We also find that models detect entrainment without relying on the same features as humans, suggesting that decisions may be driven by indirect or spurious signals. This raises concerns about robustness and reliability, as systems that appear to capture human-like interaction patterns may behave unpredictably in new settings.

To mitigate these risks, we encourage future work that prioritizes evaluation across diverse multilingual contexts. We also caution against treating entrainment as an isolated normative measure of communicative competence, and recommend safeguards such as usage restrictions, misuse monitoring, and human oversight in sensitive applications.

\section*{Acknowledgments}

We thank Nicholas Deas for several helpful discussions and feedback, and Chhavi Dixit for verifying Hindi-English LID and CSW annotations. DB is supported by the National Science Foundation under Grant IIS 2418307.

\newpage
\bibliography{custom}

@ARTICLE {masac,
author = {M. Bedi and S. Kumar and M. Akhtar and T. Chakraborty},
journal = {IEEE Transactions on Affective Computing},
title = {{Multi-modal Sarcasm Detection and Humor Classification in Code-mixed Conversations}},
issn = {1949-3045},
year={2023},
doi = {10.1109/TAFFC.2021.3083522},
publisher = {IEEE Computer Society},
address = {Los Alamitos, CA, USA},
month = {5}
}

@inproceedings{seame,
author = {Lyu, Dau-Cheng and Tan, Tien-Ping and Chng, Eng and Li, Haizhou},
year = {2010},
month = {01},
pages = {1986-1989},
title = {Mandarin–{E}nglish code-switching speech corpus in {S}outh-{E}ast {A}sia: {SEAME}},
volume = {49},
booktitle = {Language Resources and Evaluation},
doi = {10.1007/s10579-015-9303-x}
}

@misc{miami, title={Miami corpus: {P}reliminary documentation - bangortalk}, url={http://bangortalk.org.uk/docs/Miami_doc.pdf}, author={Deuchar, Margaret}, year = {2011}}

@inproceedings{bawa-etal-2018-accommodation,
    title = "{A}ccommodation of {C}onversational {C}ode-{C}hoice",
    author = "Bawa, Anshul  and
      Choudhury, Monojit  and
      Bali, Kalika",
    editor = "Aguilar, Gustavo  and
      AlGhamdi, Fahad  and
      Soto, Victor  and
      Solorio, Thamar  and
      Diab, Mona  and
      Hirschberg, Julia",
    booktitle = "Proceedings of the Third Workshop on Computational Approaches to Linguistic Code-Switching",
    month = jul,
    year = "2018",
    address = "Melbourne, Australia",
    publisher = "Association for Computational Linguistics",
    url = "https://aclanthology.org/W18-3210/",
    doi = "10.18653/v1/W18-3210",
    pages = "82--91"
}

@inproceedings{Danescu_Niculescu_Mizil_2011,
	doi = {10.1145/1963405.1963509},
	url = {https://doi.org/10.1145%2F1963405.1963509},
	year = 2011,
	month = {March},
    publisher = {{ACM}},
	author = {Cristian Danescu-Niculescu-Mizil and Michael Gamon and Susan Dumais},
	title = {{Mark my Words! Linguistic Style Accommodation in Social Media}},
	booktitle = {Proceedings of the 20th International Conference on World Wide Web}
}

@inbook{giles_coupland_coupland_1991, place={Cambridge}, series={Studies in Emotion and Social Interaction}, title={Accommodation theory: Communication, context, and consequence}, DOI={10.1017/CBO9780511663673.001}, booktitle={Contexts of Accommodation: Developments in Applied Sociolinguistics}, publisher={Cambridge University Press}, author={Giles, Howard and Coupland, Nikolas and Coupland, Justine}, year={1991}, pages={1–68}, collection={Studies in Emotion and Social Interaction}}

@article{maurer-tindall,
title = {Effect of postural congruence on client's perception of counselor empathy},
journal = {Journal of Counseling Psychology},
volume = {30(2)},
pages = {158-163},
year = {1983},
author = {Richard E. Maurer and Jeffrey H. Tindall}
}

@inproceedings{levitan11_interspeech,
  author={Rivka Levitan and Julia Hirschberg},
  title={{Measuring acoustic-prosodic entrainment with respect to multiple levels and dimensions}},
  year=2011,
  booktitle={Proc. Interspeech 2011},
  pages={3081--3084},
  doi={10.21437/Interspeech.2011-771}
}

@article{burgoon,
    author = {Burgoon, Judee K. and Buller, David B. and Hale, Jerold L. and de Turck, Mark A.},
    title = "{Relational Messages Associated with Nonverbal Behaviors}",
    journal = {Human Communication Research},
    volume = {10},
    number = {3},
    pages = {351-378},
    year = {2006},
    month = {03},
    issn = {0360-3989},
    doi = {10.1111/j.1468-2958.1984.tb00023.x},
    url = {https://doi.org/10.1111/j.1468-2958.1984.tb00023.x},
    eprint = {https://academic.oup.com/hcr/article-pdf/10/3/351/22342762/jhumcom0351.pdf},
}

@article{Chartrand1999TheCE,
  title={The chameleon effect: the perception-behavior link and social interaction.},
  author={Tanya L. Chartrand and John A. Bargh},
  journal={Journal of personality and social psychology},
  year={1999},
  volume={76 6},
  pages={
          893-910
        },
  url={https://api.semanticscholar.org/CorpusID:11818459}
}

@inproceedings{nenkova-etal-2008-high,
    title = "{High Frequency Word Entrainment in Spoken Dialogue}",
    author = "Nenkova, Ani  and
      Gravano, Agustin  and
      Hirschberg, Julia",
    booktitle = "Proceedings of ACL-08: HLT, Short Papers",
    month = jun,
    year = "2008",
    address = "Columbus, Ohio",
    publisher = "Association for Computational Linguistics",
    url = "https://aclanthology.org/P08-2043",
    pages = "169--172",
}

@inproceedings{reitter-moore-2007-predicting,
    title = "{Predicting Success in Dialogue}",
    author = "Reitter, David  and
      Moore, Johanna D.",
    booktitle = "Proceedings of the 45th Annual Meeting of the Association of Computational Linguistics",
    month = jun,
    year = "2007",
    address = "Prague, Czech Republic",
    publisher = "Association for Computational Linguistics",
    url = "https://aclanthology.org/P07-1102",
    pages = "808--815",
}

@inproceedings{levitan-etal-2012-acoustic,
    title = "{Acoustic-Prosodic Entrainment and Social Behavior}",
    author = "Levitan, Rivka  and
      Gravano, Agustin  and
      Willson, Laura  and
      Benus, Stefan  and
      Hirschberg, Julia  and
      Nenkova, Ani",
    booktitle = "Proceedings of the 2012 Conference of the North {A}merican Chapter of the Association for Computational Linguistics: Human Language Technologies",
    month = jun,
    year = "2012",
    address = "Montr{\'e}al, Canada",
    publisher = "Association for Computational Linguistics",
    url = "https://aclanthology.org/N12-1002",
    pages = "11--19",
}

@article{KOOTSTRA2010210,
title = {Syntactic alignment and shared word order in code-switched sentence production: Evidence from bilingual monologue and dialogue},
journal = {Journal of Memory and Language},
volume = {63},
number = {2},
pages = {210-231},
year = {2010},
issn = {0749-596X},
doi = {https://doi.org/10.1016/j.jml.2010.03.006},
url = {https://www.sciencedirect.com/science/article/pii/S0749596X10000288},
author = {Gerrit Jan Kootstra and Janet G. {van Hell} and Ton Dijkstra}
}

@inproceedings{ahn-etal-2020-code,
    title = "What {C}ode-{S}witching {S}trategies are {E}ffective in {D}ialog {S}ystems?",
    author = "Ahn, Emily  and
      Jimenez, Cecilia  and
      Tsvetkov, Yulia  and
      Black, Alan W",
    booktitle = "Proceedings of the Society for Computation in Linguistics 2020",
    month = jan,
    year = "2020",
    address = "New York, New York",
    publisher = "Association for Computational Linguistics",
    url = "https://aclanthology.org/2020.scil-1.32",
    pages = "254--264",
}

@inproceedings{parekh-etal-2020-understanding,
    title = "{Understanding Linguistic Accommodation in Code-Switched Human-Machine Dialogues}",
    author = "Parekh, Tanmay  and
      Ahn, Emily  and
      Tsvetkov, Yulia  and
      Black, Alan W",
    booktitle = "Proceedings of the 24th Conference on Computational Natural Language Learning",
    month = nov,
    year = "2020",
    address = "Online",
    publisher = "Association for Computational Linguistics",
    url = "https://aclanthology.org/2020.conll-1.46",
    doi = "10.18653/v1/2020.conll-1.46",
    pages = "565--577",
}

@inproceedings{bhattacharya-etal-2024-measuring,
    title = "{Measuring Entrainment in Spontaneous Code-switched Speech}",
    author = "Bhattacharya, Debasmita  and
      Ding, Siying  and
      Nguyen, Alayna  and
      Hirschberg, Julia",
    booktitle = "Proceedings of the 2024 Conference of the North American Chapter of the Association for Computational Linguistics: Human Language Technologies (Volume 1: Long Papers)",
    month = jun,
    year = "2024",
    address = "Mexico City, Mexico",
    publisher = "Association for Computational Linguistics",
    url = "https://aclanthology.org/2024.naacl-long.158/",
    doi = "10.18653/v1/2024.naacl-long.158",
    pages = "2865--2876"
}

@inproceedings{kejriwal23b_interspeech,
  title     = {{Relationship between auditory and semantic entrainment using Deep Neural Networks (DNN)}},
  author    = {Jay Kejriwal and Štefan Beňuš},
  year      = {2023},
  booktitle = {Interspeech 2023},
  pages     = {2623--2627},
  doi       = {10.21437/Interspeech.2023-1947},
  issn      = {2958-1796},
}

@inproceedings{levitan-etal-2015-acoustic,
    title = "Acoustic-prosodic entrainment in {S}lovak, {S}panish, {E}nglish and {C}hinese: A cross-linguistic comparison",
    author = "Levitan, Rivka  and
      Be{\v{n}}u{\v{s}}, {\v{S}}tefan  and
      Gravano, Agust{\'i}n  and
      Hirschberg, Julia",
    editor = "Koller, Alexander  and
      Skantze, Gabriel  and
      Jurcicek, Filip  and
      Araki, Masahiro  and
      Rose, Carolyn Penstein",
    booktitle = "Proceedings of the 16th Annual Meeting of the Special Interest Group on Discourse and Dialogue",
    month = sep,
    year = "2015",
    address = "Prague, Czech Republic",
    publisher = "Association for Computational Linguistics",
    url = "https://aclanthology.org/W15-4644/",
    doi = "10.18653/v1/W15-4644",
    pages = "325--334"
}

@inproceedings{xia14_speechprosody,
  title     = {{Prosodic Entrainment in Mandarin Chinese and English: A Cross-Linguistic Comparison}},
  author    = {Zhihua Xia and Rivka Levitan and Julia Hirschberg},
  year      = {2014},
  booktitle = {{Speech Prosody 2014}},
  pages     = {65--69},
  doi       = {10.21437/SpeechProsody.2014-1},
  issn      = {2333-2042},
}

@inproceedings{weise-etal-2021-talk,
    title = "``{T}alk to me with left, right, and angles'': Lexical entrainment in spoken {H}ebrew dialogue",
    author = "Weise, Andreas  and
      Silber-Varod, Vered  and
      Lerner, Anat  and
      Hirschberg, Julia  and
      Levitan, Rivka",
    editor = "Merlo, Paola  and
      Tiedemann, Jorg  and
      Tsarfaty, Reut",
    booktitle = "Proceedings of the 16th Conference of the European Chapter of the Association for Computational Linguistics: Main Volume",
    month = apr,
    year = "2021",
    address = "Online",
    publisher = "Association for Computational Linguistics",
    url = "https://aclanthology.org/2021.eacl-main.23/",
    doi = "10.18653/v1/2021.eacl-main.23",
    pages = "292--299"
}

@inproceedings{soto18_interspeech,
  title     = {{The Role of Cognate Words, POS Tags and Entrainment in Code-Switching}},
  author    = {Victor Soto and Nishmar Cestero and Julia Hirschberg},
  year      = {2018},
  booktitle = {{Interspeech 2018}},
  pages     = {1938--1942},
  doi       = {10.21437/Interspeech.2018-1099},
  issn      = {2958-1796},
}

@INPROCEEDINGS{gravano_tobi,
  author={Gravano, Agustín and Beňuš, Štefan and Levitan, Rivka and Hirschberg, Julia},
  booktitle={2014 IEEE Spoken Language Technology Workshop (SLT)}, 
  title={Three {ToBI}-based measures of prosodic entrainment and their correlations with speaker engagement}, 
  year={2014},
  volume={},
  number={},
  pages={578-583},
  doi={10.1109/SLT.2014.7078638}}

@inproceedings{havaldar-etal-2023-comparing,
    title = "{Comparing Styles across Languages}",
    author = "Havaldar, Shreya  and
      Pressimone, Matthew  and
      Wong, Eric  and
      Ungar, Lyle",
    editor = "Bouamor, Houda  and
      Pino, Juan  and
      Bali, Kalika",
    booktitle = "Proceedings of the 2023 Conference on Empirical Methods in Natural Language Processing",
    month = dec,
    year = "2023",
    address = "Singapore",
    publisher = "Association for Computational Linguistics",
    url = "https://aclanthology.org/2023.emnlp-main.419/",
    doi = "10.18653/v1/2023.emnlp-main.419",
    pages = "6775--6791"
}

@inproceedings{bhattacharya-etal-2025-code,
    title = "{Code-switching in Context: Investigating the Role of Discourse Topic in Bilingual Speech Production}",
    author = "Bhattacharya, Debasmita  and
      Yi, Anxin  and
      Ding, Siying  and
      Hirschberg, Julia",
    editor = "Strube, Michael  and
      Braud, Chloe  and
      Hardmeier, Christian  and
      Li, Junyi Jessy  and
      Loaiciga, Sharid  and
      Zeldes, Amir  and
      Li, Chuyuan",
    booktitle = "Proceedings of the 6th Workshop on Computational Approaches to Discourse, Context and Document-Level Inferences (CODI 2025)",
    month = nov,
    year = "2025",
    address = "Suzhou, China",
    publisher = "Association for Computational Linguistics",
    url = "https://aclanthology.org/2025.codi-1.6/",
    doi = "10.18653/v1/2025.codi-1.6",
    pages = "64--80",
    ISBN = "979-8-89176-343-2"
}

@misc{ethnologue,
  title        = {What is the most spoken language?},
  author       = {Ethnologue},
  year         = 2026,
  howpublished = {\url{https://www.ethnologue.com/insights/most-spoken-language/}},
  note         = {Accessed: 2026-04-06}
}

@inproceedings{namazifard-poech-2025-isolating,
    title = "{Isolating Culture Neurons in Multilingual Large Language Models}",
    author = "Namazifard, Danial  and
      Galke Poech, Lukas",
    editor = "Inui, Kentaro  and
      Sakti, Sakriani  and
      Wang, Haofen  and
      Wong, Derek F.  and
      Bhattacharyya, Pushpak  and
      Banerjee, Biplab  and
      Ekbal, Asif  and
      Chakraborty, Tanmoy  and
      Singh, Dhirendra Pratap",
    booktitle = "Proceedings of the 14th International Joint Conference on Natural Language Processing and the 4th Conference of the Asia-Pacific Chapter of the Association for Computational Linguistics",
    month = dec,
    year = "2025",
    address = "Mumbai, India",
    publisher = "The Asian Federation of Natural Language Processing and The Association for Computational Linguistics",
    url = "https://aclanthology.org/2025.findings-ijcnlp.45/",
    doi = "10.18653/v1/2025.findings-ijcnlp.45",
    pages = "768--785",
    ISBN = "979-8-89176-303-6"
}

@inproceedings{blevins-etal-2026-language,
    title = "Do language models accommodate their users? {A} study of linguistic convergence",
    author = "Blevins, Terra  and
      Schmalwieser, Susanne  and
      Roth, Benjamin",
    editor = "Demberg, Vera  and
      Inui, Kentaro  and
      Marquez, Llu{\'i}s",
    booktitle = "Proceedings of the 19th Conference of the {E}uropean Chapter of the {A}ssociation for {C}omputational {L}inguistics (Volume 1: Long Papers)",
    month = mar,
    year = "2026",
    address = "Rabat, Morocco",
    publisher = "Association for Computational Linguistics",
    url = "https://aclanthology.org/2026.eacl-long.34/",
    doi = "10.18653/v1/2026.eacl-long.34",
    pages = "791--807",
    ISBN = "979-8-89176-380-7"
}

@inproceedings{lin-etal-2025-v-vae,
    title = "{V}-{VAE}: {A Variational Auto Encoding Framework Towards Fine-Grained Control over Human-Like Chat}",
    author = "Lin, Qi  and
      Xu, Weikai  and
      Chen, Lisi  and
      Dai, Bin",
    editor = "Christodoulopoulos, Christos  and
      Chakraborty, Tanmoy  and
      Rose, Carolyn  and
      Peng, Violet",
    booktitle = "Proceedings of the 2025 Conference on Empirical Methods in Natural Language Processing",
    month = nov,
    year = "2025",
    address = "Suzhou, China",
    publisher = "Association for Computational Linguistics",
    url = "https://aclanthology.org/2025.emnlp-main.1508/",
    doi = "10.18653/v1/2025.emnlp-main.1508",
    pages = "29693--29706",
    ISBN = "979-8-89176-332-6"
}

@article{parselmouth,
    author = "Yannick Jadoul and Bill Thompson and Bart de Boer",
    title = "Introducing {P}arselmouth: A {P}ython interface to {P}raat",
    journal = "Journal of Phonetics",
    volume = "71",
    pages = "1--15",
    year = "2018",
    doi = "https://doi.org/10.1016/j.wocn.2018.07.001"
}

@book{muysken-2000,
  title={Bilingual speech: a typology of code-mixing},
  author={Muysken, Pieter},
  year={2000},
  location={Cambridge, UK},
  publisher={Cambridge University Press}
}

@article{poplack-1980,
author = {Poplack, Shana},
year = {1980},
month = {01},
pages = {581-618},
title = {Sometimes {I}'ll start a sentence in {S}panish {Y TERMINO EN ESPAÑOL}: toward a typology of code-switching},
volume = {18},
journal = {Linguistics},
doi = {10.1515/ling.1980.18.7-8.581}
}

@misc{huang2020tabtransformertabulardatamodeling,
      title={{TabTransformer: Tabular Data Modeling Using Contextual Embeddings}}, 
      author={Xin Huang and Ashish Khetan and Milan Cvitkovic and Zohar Karnin},
      year={2020},
      eprint={2012.06678},
      archivePrefix={arXiv},
      primaryClass={cs.LG},
      url={https://arxiv.org/abs/2012.06678}, 
}

@inproceedings{ft-transformer,
author = {Dai, Huangliang and Wu, Shixun and Huang, Jiajun and Jian, Zizhe and Zhu, Yue and Hu, Haiyang and Chen, Zizhong},
title = {{FT-Transformer: Resilient and Reliable Transformer with End-to-End Fault Tolerant Attention}},
year = {2025},
isbn = {9798400714665},
publisher = {Association for Computing Machinery},
address = {New York, NY, USA},
url = {https://doi.org/10.1145/3712285.3759853},
doi = {10.1145/3712285.3759853},
booktitle = {Proceedings of the International Conference for High Performance Computing, Networking, Storage and Analysis},
pages = {1085–1098},
numpages = {14},
location = {
},
series = {SC '25}
}

@inproceedings{xlm-roberta,
    title = "{Unsupervised Cross-lingual Representation Learning at Scale}",
    author = "Conneau, Alexis  and
      Khandelwal, Kartikay  and
      Goyal, Naman  and
      Chaudhary, Vishrav  and
      Wenzek, Guillaume  and
      Guzm{\'a}n, Francisco  and
      Grave, Edouard  and
      Ott, Myle  and
      Zettlemoyer, Luke  and
      Stoyanov, Veselin",
    editor = "Jurafsky, Dan  and
      Chai, Joyce  and
      Schluter, Natalie  and
      Tetreault, Joel",
    booktitle = "Proceedings of the 58th Annual Meeting of the Association for Computational Linguistics",
    month = jul,
    year = "2020",
    address = "Online",
    publisher = "Association for Computational Linguistics",
    url = "https://aclanthology.org/2020.acl-main.747/",
    doi = "10.18653/v1/2020.acl-main.747",
    pages = "8440--8451"
}

@misc{devlin2019bertpretrainingdeepbidirectional,
      title={{BERT}: Pre-training of {D}eep {B}idirectional {T}ransformers for {L}anguage {U}nderstanding}, 
      author={Jacob Devlin and Ming-Wei Chang and Kenton Lee and Kristina Toutanova},
      year={2019},
      eprint={1810.04805},
      archivePrefix={arXiv},
      primaryClass={cs.CL},
      url={https://arxiv.org/abs/1810.04805}, 
}

@misc{chung2020rethinkingembeddingcouplingpretrained,
      title={Rethinking embedding coupling in pre-trained language models}, 
      author={Hyung Won Chung and Thibault Févry and Henry Tsai and Melvin Johnson and Sebastian Ruder},
      year={2020},
      eprint={2010.12821},
      archivePrefix={arXiv},
      primaryClass={cs.CL},
      url={https://arxiv.org/abs/2010.12821}, 
}

@article{Chen2021WavLMLS,
  title={{WavLM: Large-Scale Self-Supervised Pre-Training for Full Stack Speech Processing}},
  author={Sanyuan Chen and Chengyi Wang and Zhengyang Chen and Yu Wu and Shujie Liu and Zhuo Chen and Jinyu Li and Naoyuki Kanda and Takuya Yoshioka and Xiong Xiao and Jian Wu and Long Zhou and Shuo Ren and Yanmin Qian and Yao Qian and Micheal Zeng and Furu Wei},
  journal={IEEE Journal of Selected Topics in Signal Processing},
  year={2021},
  volume={16},
  pages={1505-1518},
  url={https://api.semanticscholar.org/CorpusID:239885872}
}

@inproceedings{zanonboito24_interspeech,
  title     = {{mHuBERT-147: A Compact Multilingual HuBERT Model}},
  author    = {Marcely {Zanon Boito} and Vivek Iyer and Nikolaos Lagos and Laurent Besacier and Ioan Calapodescu},
  year      = {2024},
  booktitle = {{Interspeech 2024}},
  pages     = {3939--3943},
  doi       = {10.21437/Interspeech.2024-938},
  issn      = {2958-1796},
}

@inproceedings{SHAP,
author = {Lundberg, Scott M. and Lee, Su-In},
title = {A unified approach to interpreting model predictions},
year = {2017},
isbn = {9781510860964},
publisher = {Curran Associates Inc.},
address = {Red Hook, NY, USA},
booktitle = {Proceedings of the 31st International Conference on Neural Information Processing Systems},
pages = {4768–4777},
numpages = {10},
location = {Long Beach, California, USA},
series = {NIPS'17}
}

@article{m-index,

title ={The {LIDES} {C}oding {M}anual: {A} document for preparing and analyzing language interaction data Version 1.1—{J}uly, 1999.},
author = "Barnett, Ruthanna  and
      Codó, Eva   and
      Eppler, Eva  and
      Forcadell, Montse and
Gardner-Chloros, Penelope and
van Hout, Roeland  and
Moyer, Melissa and
Torras, Maria Carme and
Turell, Maria Teresa and
Sebba, Mark and
Starren, Marianne and
Wensing, Sietse",
journal = {International Journal of Bilingualism},
volume = {4},
number = {2},
pages = {131-270},
year = {2000},
doi = {10.1177/13670069000040020107},
URL = {https://doi.org/10.1177/13670069000040020107
},
eprint = {    https://doi.org/10.1177/13670069000040020107
}
}

@inproceedings{i-index,
    title = "{Simple Tools for Exploring Variation in Code-switching for Linguists}",
    author = "Guzman, Gualberto A.  and
      Serigos, Jacqueline  and
      Bullock, Barbara E.  and
      Toribio, Almeida Jacqueline",
    editor = "Diab, Mona  and
      Fung, Pascale  and
      Ghoneim, Mahmoud  and
      Hirschberg, Julia  and
      Solorio, Thamar",
    booktitle = "Proceedings of the Second Workshop on Computational Approaches to Code Switching",
    month = nov,
    year = "2016",
    address = "Austin, Texas",
    publisher = "Association for Computational Linguistics",
    url = "https://aclanthology.org/W16-5802/",
    doi = "10.18653/v1/W16-5802",
    pages = "12--20"
}

@inproceedings{briakou-etal-2021-ola,
    title = "Ol{\'a}, Bonjour, Salve! {XFORMAL}: {A} {B}enchmark for {M}ultilingual {F}ormality {S}tyle {T}ransfer",
    author = "Briakou, Eleftheria  and
      Lu, Di  and
      Zhang, Ke  and
      Tetreault, Joel",
    editor = "Toutanova, Kristina  and
      Rumshisky, Anna  and
      Zettlemoyer, Luke  and
      Hakkani-Tur, Dilek  and
      Beltagy, Iz  and
      Bethard, Steven  and
      Cotterell, Ryan  and
      Chakraborty, Tanmoy  and
      Zhou, Yichao",
    booktitle = "Proceedings of the 2021 Conference of the North American Chapter of the Association for Computational Linguistics: Human Language Technologies",
    month = jun,
    year = "2021",
    address = "Online",
    publisher = "Association for Computational Linguistics",
    url = "https://aclanthology.org/2021.naacl-main.256/",
    doi = "10.18653/v1/2021.naacl-main.256",
    pages = "3199--3216"
}

@inproceedings{srinivasan-choi-2022-tydip,
    title = "{T}y{D}i{P}: {A Dataset for Politeness Classification in Nine Typologically Diverse Languages}",
    author = "Srinivasan, Anirudh  and
      Choi, Eunsol",
    editor = "Goldberg, Yoav  and
      Kozareva, Zornitsa  and
      Zhang, Yue",
    booktitle = "Findings of the Association for Computational Linguistics: EMNLP 2022",
    month = dec,
    year = "2022",
    address = "Abu Dhabi, United Arab Emirates",
    publisher = "Association for Computational Linguistics",
    url = "https://aclanthology.org/2022.findings-emnlp.420/",
    doi = "10.18653/v1/2022.findings-emnlp.420",
    pages = "5723--5738"
}

@inproceedings{Hua2025:post-train,
  title     = {{Post-training for Efficient Communication via Convention Formation}},
  author    = {Hua, Yilun and Wang, Evan and Artzi, Yoav},
  booktitle = {Proceedings of the Conference on Language Modeling},
  year      = {2025},
  url       = {https://arxiv.org/abs/2508.06482}
}

@inproceedings{
Dathathri2020Plug,
title={{Plug and Play Language Models: A Simple Approach to Controlled Text Generation}},
author={Sumanth Dathathri and Andrea Madotto and Janice Lan and Jane Hung and Eric Frank and Piero Molino and Jason Yosinski and Rosanne Liu},
booktitle={International Conference on Learning Representations},
year={2020},
url={https://openreview.net/forum?id=H1edEyBKDS}
}

@inproceedings{scalena-etal-2026-steering,
    title = "{Steering Large Language Models for Machine Translation Personalization}",
    author = "Scalena, Daniel  and
      Sarti, Gabriele  and
      Bisazza, Arianna  and
      Fersini, Elisabetta  and
      Nissim, Malvina",
    editor = "Demberg, Vera  and
      Inui, Kentaro  and
      Marquez, Llu{\'i}s",
    booktitle = "Proceedings of the 19th Conference of the {E}uropean Chapter of the {A}ssociation for {C}omputational {L}inguistics (Volume 1: Long Papers)",
    month = mar,
    year = "2026",
    address = "Rabat, Morocco",
    publisher = "Association for Computational Linguistics",
    url = "https://aclanthology.org/2026.eacl-long.217/",
    doi = "10.18653/v1/2026.eacl-long.217",
    pages = "4681--4701",
    ISBN = "979-8-89176-380-7"
}

@inproceedings{dinan-etal-2020-multi,
    title = "{Multi-Dimensional Gender Bias Classification}",
    author = "Dinan, Emily  and
      Fan, Angela  and
      Wu, Ledell  and
      Weston, Jason  and
      Kiela, Douwe  and
      Williams, Adina",
    editor = "Webber, Bonnie  and
      Cohn, Trevor  and
      He, Yulan  and
      Liu, Yang",
    booktitle = "Proceedings of the 2020 Conference on Empirical Methods in Natural Language Processing (EMNLP)",
    month = nov,
    year = "2020",
    address = "Online",
    publisher = "Association for Computational Linguistics",
    url = "https://aclanthology.org/2020.emnlp-main.23/",
    doi = "10.18653/v1/2020.emnlp-main.23",
    pages = "314--331"
}

@inproceedings{liang-etal-2024-controlled,
    title = "{Controlled Text Generation for Large Language Model with Dynamic Attribute Graphs}",
    author = "Liang, Xun  and
      Wang, Hanyu  and
      Song, Shichao  and
      Hu, Mengting  and
      Wang, Xunzhi  and
      Li, Zhiyu  and
      Xiong, Feiyu  and
      Tang, Bo",
    editor = "Ku, Lun-Wei  and
      Martins, Andre  and
      Srikumar, Vivek",
    booktitle = "Findings of the Association for Computational Linguistics: ACL 2024",
    month = aug,
    year = "2024",
    address = "Bangkok, Thailand",
    publisher = "Association for Computational Linguistics",
    url = "https://aclanthology.org/2024.findings-acl.345/",
    doi = "10.18653/v1/2024.findings-acl.345",
    pages = "5797--5814"
}

@inproceedings{danescu-niculescu-mizil-lee-2011-chameleons,
    title = "{Chameleons in Imagined Conversations: A New Approach to Understanding Coordination of Linguistic Style in Dialogs}",
    author = "Danescu-Niculescu-Mizil, Cristian  and
      Lee, Lillian",
    editor = "Keller, Frank  and
      Reitter, David",
    booktitle = "Proceedings of the 2nd Workshop on Cognitive Modeling and Computational Linguistics",
    month = jun,
    year = "2011",
    address = "Portland, Oregon, USA",
    publisher = "Association for Computational Linguistics",
    url = "https://aclanthology.org/W11-0609/",
    pages = "76--87"
}

\newpage
\appendix

\section{Appendix}
\label{sec:appendix}

\subsection{Computing code-switched entrainment: definitions \& full methodological details.}
\label{sec:app_method}

We reproduce the method of \citet{bhattacharya-etal-2024-measuring} here for completeness. Note that \textit{proximity} refers to absolute similarity in speech, \textit{convergence} refers to how much interlocutors' speech becomes more similar over time, and \textit{synchrony} refers to relative coordination between interlocutors' speech. 


We compute \textbf{lexical entrainment} over word classes using Eq.~\ref{eq1}  \cite{nenkova-etal-2008-high}, which
calculates entrainment between speakers $S_A$ and $S_B$ on a given word class $W$ as the negated sum of absolute differences in each word $w \in W$'s usage rate between $S_A$ and $S_B$. 

\texttt{\small{\begin{equation}
entr(S_A, S_B)=-\sum_{w \in W}|\frac{count_{S_A}(w)}{ALL_{S_A}}-\frac{count_{S_B}(w)}{ALL_{S_B}}|
\label{eq1}
\end{equation}}}

We also compute \textbf{lexical entrainment} on overall language use in each corpus using Kneser-Essen-Ney smoothed trigram language models trained on each speaker's transcripts via the KenLM toolkit. For each conversation, we evaluate the model trained on $S_A$'s transcripts through its perplexity on $S_B$'s speech transcripts. We do the opposite to evaluate the $S_B$-trained model. Resulting negated perplexity values are unidirectional entrainment scores; lower perplexity reflects greater entrainment to the interlocutor.

We study entrainment over \textbf{acoustic-prosodic} and \textbf{CSW style} features via turn- and conversation-level proximity, convergence, and synchrony. 

At the \textbf{turn-level}, we measure \textbf{proximity} by computing a \textit{partner difference} (Eq.~\ref{eq2}) and \textit{other difference} (Eq.~\ref{eq3}) for each target speaker turn, such that $turn_p$ is adjacent to the target turn and uttered by the target turn speaker’s conversation partner, and $turn_i$ is uttered by the target turn speaker’s conversation partner but is not adjacent to the target turn, over ten random turns. We compare \textit{partner differences} and \textit{other differences} for a given feature with a paired \textit{t}-test, and infer proximity when partner differences from the prior speaker turn are smaller than differences from other speaker turns.

\texttt{\small{\begin{equation}
    difference_{partner} = |turn_t - turn_p|
\label{eq2}
\end{equation}}}
\texttt{\small{\begin{equation}
    difference_{other} = \frac{\sum_{i=1}^{10}|turn_t - turn_i|}{10}
\label{eq3}
\end{equation}}}

For turn-level \textbf{convergence}, we compute the Pearson correlation coefficient between the absolute feature difference between adjacent turns and the turn number. We infer strong, moderate, and weak convergence from $r \geq 0.7$, $0.5 \leq r<0.7$, and $0<r<0.5$, respectively. We similarly infer strong, moderate, and weak divergence from $r \leq -0.7$, $-0.7<r \leq -0.5$, and $-0.5<r<0$. We similarly compute turn-level \textbf{synchrony}, using Pearson correlation between feature values of adjacent turns from different speakers, testing for significance with a two-sided $t$-test. We infer the degree of synchrony or asynchrony using the same ranges of $r$-values as for turn-level convergence. 

At the \textbf{conversation-level}, we calculate \textbf{proximity} using paired $t$-tests to compare two sets of differences for each speaker: \textit{partner differences}, i.e., the difference between the speaker's value for a given feature and that of their partner, versus \textit{other differences}, i.e., the mean of differences between the speaker’s value and the values of each speaker who was not their interlocutor. We infer proximity when partner differences are smaller than other differences. We compute \textbf{convergence} by halving each conversation and comparing the difference in the two speakers' mean values for a given feature in the first half to that in the second half using a paired $t$-test. We infer convergence when differences in the second half are significantly smaller.

\subsection{Datasets' license information.}
\label{sec:app_datalicenses}
The SEAME corpus may be used for research per the \href{https://catalog.ldc.upenn.edu/license/ldc-non-members-agreement.pdf}{LDC User Agreement for Non-Members}. The MaSaC corpus is released by the authors in a public Github repository at \url{https://github.com/LCS2-IIITD/MSH-COMICS}. The BM corpus is made available under the \href{http://gnu.org/copyleft/gpl.html}{GNU General Public
License} version 3 or later.  Our use of these corpora is consistent with their intended use cases in research.

\subsection{Affirmative cues and fillers.}
\label{sec:app_wordclasses} 

The members of the \textit{affirmative cues} word class for SEAME are: `yeah', `right', `ok', `mmhmm', `ya', `yes', `huh', `alright', `yup', `hah', `yep', `yah', `\zh{对}', `\zh{嗯}', `\zh{啊哈}', `\zh{啊哈}', `\zh{哦}', `\zh{哈}', `\zh{是}', `\zh{当然}', `\zh{可以}', `\zh{行}'. We derive this list through a combination of SEAME data exploration and reference to the following articles: 

\begin{itemize}
    \item \small \url{https://direct.mit.edu/coli/article/38/1/1/2144/Affirmative-Cue-Words-in-Task-Oriented-Dialogue}
    \item \url{https://lcchineseschool.com/say-yes-in-chinese-quick-guide-to-affirmation/}
\end{itemize}

The members of the \textit{fillers} word class for SEAME are: `um', `ah', `uh', `eh', `er', `hmm', `mm', `mmm', `umm', `ar', `hm',
           `\zh{啊}', `\zh{呃}', `\zh{呃}', `\zh{啊}', `\zh{呃}', `\zh{嗯}', `\zh{嗯}', `\zh{哼}'. This list is derived similarly to the one for affirmative cues in SEAME, with additional reference to the article at \url{https://www.chineseclass101.com/blog/2021/09/09/chinese-filler-words/}.

The members of the \textit{affirmative cues} word class for MaSaC are: `haan', `arey', `aray', `arre', `oh', `ooh', `ohh', `achchha', `achcha', `ya', `yah', `vah', `wah', `waah', `ok',
                         `okay', `ha', `bol', `right', `oho', `oo', `huh', `yes', `alright', `hein', `really', `jee', `waa'.

The members of the \textit{fillers} word class for MaSaC are: `voh', `woh', `wooh', `aa', `aah', `ahh' `aaah', `na', `umm', `hmm', `ae', `achchha', `achcha', `pch', `aaa', `mmm', `o', `im',
           `oye', `haww', `uhhh', `uhh', `m', `ahem', `mmmmm', `uh'. 

Both word classes for MaSaC are derived through data exploration. 

\subsection{Pre-trained models' license information.}
\label{sec:app_modellicenses}

Our use of these models is consistent with their intended use cases in research. 
\begin{itemize}
    \item \textit{TabTransformer}: Creative Commons CC0 1.0 Universal.
    \item \textit{FT-Transformer}: Creative Commons Attribution 4.0 International.
    \item \textit{XLM-R}: MIT license. 
    \item \textit{mBERT}: Apache License 2.0
    \item \textit{RemBERT}: Apache License 2.0. 
    \item \textit{WaVLM}: Creative Commons Attribution-ShareAlike 3.0 Unported
    \item \textit{mHuBERT}: Creative Commons Attribution Non Commercial Share Alike 4.0
\end{itemize}

\subsection{Hyperparameters for grid search for each model.}
\label{sec:app_hyperparams}

Within Table~\ref{tab:hyperparams}, all models from \textit{TabTransformer} onwards encode input data with the stated architecture, which is then fused with an MLP and softmax function for predictions. We encode transcript text alongside tabular data as input in the \textit{XLM-R}, \textit{mBERT}, and \textit{RemBERT} models, splitting transcripts into 512-token chunks using the corresponding tokenizer for each model. The parameters of each of these pre-trained encoders are frozen so that only the numerical \textit{MLP}, pooling, fusion, and classifier train, and feature importance attribution 
is computed only over tabular features. We encode audio recordings alongside tabular data as input in the \textit{WavLM} and \textit{mHuBERT} models, chunking audio in 5-second segments and using averaged pooled acoustic-prosodic embeddings for each. We similarly freeze the parameters of these encoders during training. For each corpus, the final multimodal model combining text and speech embeddings with tabular inputs is determined based on the best-performing text and speech embedding methods (\textit{mBERT} + \textit{mHuBERT} for BM, \textit{RemBERT} + \textit{WavLM} for SEAME, \textit{mBERT} + \textit{WavLM} for MaSaC). All models use an Adam optimizer and binary cross-entropy loss over 10 epochs and batch size of 32. Note that we adapt the \textit{TabTransformer} architecture since our data lacks categorical features after pre-processing for evaluation. 

Conversation-level experiments collectively take about 130 hours to run on a Google Colab A100 GPU. Turn-level experiments collectively take about 
460 hours to run on an A5500 GPU.

\begin{table*}[t]
\scriptsize
\centering
\renewcommand{\arraystretch}{0.95}

\begin{tabular}{
>{\raggedright\arraybackslash}p{4.2cm}
>{\centering\arraybackslash}p{1.3cm}
>{\raggedright\arraybackslash}p{8.5cm}
}
\toprule
\textbf{Model} & \textbf{\#Parameters} & \textbf{Hyperparameters tried} \\
\midrule

\texttt{LogisticRegression(class\_weight='balanced', max\_iter=1000)} & 43 &
\textbullet\ \texttt{C = [0.001, 0.001, 0.1, 0.25, 0.5, 0.75, 1, 1.25, 1.5, 1.75, 2, 10]} \\

\midrule

\texttt{SVC(class\_weight='balanced', kernel='linear', max\_iter=-1)} & 43 &
\textbullet\ \texttt{C = [0.001, 0.001, 0.1, 0.25, 0.5, 0.75, 1, 1.25, 1.5, 1.75, 2, 10]} \\

\midrule

\texttt{DecisionTreeClassifier(class\_weight='balanced')} & 22 / 3K / 98K$^{a}$ &
\textbullet\ \texttt{criterion = ['gini', 'entropy', 'log\_loss']}\par
\textbullet\ \texttt{max\_depth = [1, 3, 5, 10, 12, 15, 20, 25, 30]}\par
\textbullet\ \texttt{min\_samples\_split = [2, 5, 10]} \\

\midrule

\texttt{RandomForestClassifier(class\_weight='balanced')} & 400 / 9.4K$^{a}$ & 
\textbullet\ \texttt{criterion = ['gini', 'entropy', 'log\_loss']}\par
\textbullet\ \texttt{max\_depth = [1, 3, 5, 10, 12, 15, 20, 25, 30]}\par
\textbullet\ \texttt{min\_samples\_split = [2, 5, 10]} \\

\midrule

\texttt{GaussianNB()} & 170 & NA \\

\midrule

\texttt{KNeighborsClassifier()} & 0 &
\textbullet\ \texttt{n\_neighbors = [1, 3, 5, \dots, 49]}\par
\textbullet\ \texttt{weights = ['distance', 'uniform']}\par
\textbullet\ \texttt{algorithm = ['ball\_tree', 'kd\_tree', 'brute']}\par
\textbullet\ \texttt{leaf\_size = [5, 10, \dots, 45]} \\

\midrule

\texttt{MLPClassifier(max\_iter=1000)} & 4.5K & 
\textbullet\ \texttt{hidden\_layer\_sizes = [10, 50, 100, \dots, 500]}\par
\textbullet\ \texttt{activation = ['identity', 'logistic', 'tanh', 'relu']}\par
\textbullet\ \texttt{solver = ['lbfgs', 'adam']}\par
\textbullet\ \texttt{alpha = [1e-5, 1e-4, 1e-3, 1e-2, 1e-1, 1]} \\

\midrule

\texttt{GradientBoostingClassifier()} & 1.1K / 3.3K / 7.7K$^{b}$ & 
\textbullet\ \texttt{loss = ['log\_loss', 'exponential']}\par
\textbullet\ \texttt{learning\_rate = [1e-4, 1e-3, 0.01, 0.05, 0.1, 0.25, 0.5, 1]}\par
\textbullet\ \texttt{n\_estimators = [50, 100, \dots, 500]}\par
\textbullet\ \texttt{criterion = ['friedman\_mse', 'squared\_error']} \\

\midrule

\texttt{XGBClassifier(scale\_pos\_weight=pos\_class\_weight, max\_delta\_step=1)} & 1.1K / 1.2K / 11K$^{c}$ & 
\textbullet\ \texttt{learning\_rate = [1e-4, 1e-3, 0.01, 0.05, 0.1, 0.25, 0.5, 1]}\par
\textbullet\ \texttt{n\_estimators = [50, 100, \dots, 500]}\par
\textbullet\ \texttt{max\_depth = [1, 3, 5, 10]}\par 
\textbullet\ \texttt{booster = ['gbtree']}\\

\midrule

\texttt{(adapted) TabTransformer(emb\_dim=16, transformer\_dim=32, depth=2, heads=4) } & 141K & 
\textbullet\ \texttt{learning\_rate = [5e-5, 2e-5, 1e-5, 1e-4, 1e-3, 0.01, 0.05, 0.1, 0.25, 0.5, 1]}\\

\midrule

\texttt{FT-Transformer(emb\_dim=32, depth=3, heads=4, dropout=0.1)} & 43K & 
\textbullet\ \texttt{learning\_rate = [5e-5, 2e-5, 1e-5, 1e-4, 1e-3, 0.01, 0.05, 0.1, 0.25, 0.5, 1]}\\

\midrule

\texttt{XLM-R(hidden\_dim=128)} & 133K (+278M frozen) & 
\textbullet\ \texttt{learning\_rate = [5e-5, 2e-5, 1e-5, 1e-4, 1e-3, 0.01, 0.05, 0.1, 0.25, 0.5, 1]}\\

\midrule

\texttt{mBERT(hidden\_dim=128)} & 133K (+178M frozen) & 
\textbullet\ \texttt{learning\_rate = [5e-5, 2e-5, 1e-5, 1e-4, 1e-3, 0.01, 0.05, 0.1, 0.25, 0.5, 1]}\\

\midrule

\texttt{RemBERT(hidden\_dim=128)} & 182K (+575M frozen) & 
\textbullet\ \texttt{learning\_rate = [5e-5, 2e-5, 1e-5, 1e-4, 1e-3, 0.01, 0.05, 0.1, 0.25, 0.5, 1]}\\

\midrule

\texttt{WavLM-Base-Plus(sample\_rate=16000, emb\_dim=768, hidden\_dim=256, dropout=0.2)} & 986K & 
\textbullet\ \texttt{learning\_rate = [5e-5, 2e-5, 1e-5, 1e-4, 1e-3, 0.01, 0.05, 0.1, 0.25, 0.5, 1]}\\

\midrule

\texttt{mHuBERT-147(sample\_rate=16000, emb\_dim=768, hidden\_dim=256, dropout=0.2)} & 986K & 
\textbullet\ \texttt{learning\_rate = [5e-5, 2e-5, 1e-5, 1e-4, 1e-3, 0.01, 0.05, 0.1, 0.25, 0.5, 1]}\\

\midrule

\texttt{mBERT/RembERT + WavLM/mHuBERT(hidden\_dim=256, dropout=0.2)} & 724K (+178M frozen) / 822K (+575M frozen)$^{d}$ & 
\textbullet\ \texttt{learning\_rate = [5e-5, 2e-5, 1e-5, 1e-4, 1e-3, 0.01, 0.05, 0.1, 0.25, 0.5, 1]}\\

\bottomrule
\end{tabular}

\caption{Models with number of learned parameters and hyperparameters tried in grid search. 
\textbf{Note:} $^{a}$ depends on \texttt{max\_depth};
$^{b}$ depends on \texttt{n\_estimators};
$^{c}$ depends on \texttt{max\_depth} and \texttt{n\_estimators};
$^{d}$ depends on the model combination selected.}
\label{tab:hyperparams}
\end{table*}

\subsection{Evaluating model entrainment behavior: data construction details.}
\label{sec:app_posnegclasses}

At the \textbf{conversation-level}, we identify conversations that exhibit significant proximity on all CSW style metrics (for BM and SEAME) or lexical word classes (for MaSaC); positive (entraining) and negative (non-entraining) classes per corpus are roughly balanced and directly informed by the statistical results in Section~\ref{sec:part1_results}. We structure model inputs as a feature vector for each conversation covering speaker-separated lexical, acoustic-prosodic, and CSW style attributes, as defined in Section~\ref{sec:method_stats}; these hand-selected features effectively approximate an entrainment lexicon. We use the conversation-level mean of continuous features and convert utterance-level non-continuous features to binary flags (similar to one-hot encodings) before computing their conversation-level mean within each feature vector. 
This results in a \textit{tabular}/\textit{numerical} representation, which we concatenate with model-specific embeddings of conversation transcripts (text embeddings) or audio recordings (speech embeddings). 

At the \textbf{turn-level}, 
we apply Section~\ref{sec:part1_results}'s  statistical results to identify 
pairs of consecutive turns that exhibit significant proximity 
on a majority of CSW style metrics (for BM and SEAME) or acoustic-prosodic features (for MaSaC), handling class imbalance with oversampling. Turn-level feature vectors have the same speaker-separated structure and treatment of continuous vs. non-continuous features for associated turn pairs as do conversation-level ones; since a single speaker's turn may comprise multiple consecutive utterances, we use the mean of utterance-level values within that turn in the 
corresponding feature vector. 

We acknowledge one apparent tension in our design: at both levels of conversational granularity, the positive class is defined by significant entrainment on a particular feature set (e.g., significant proximity on CSW style metrics for conversation-level BM), and we evaluate whether models assign high importance to that same feature set. In principle, a model that perfectly learns to perform the classification task as defined should rely on those features. Our findings in Section~\ref{sec:part2_results} on imperfect alignment, where classifiers perform reasonably while bypassing the defining feature set, is therefore more striking than it may first appear; it suggests that classifiers may find alternative shortcuts rather than solving the problem as defined. We note two important caveats to this interpretation, however. First, classifiers could, in principle, achieve high accuracy via unmeasured causal features outside our theoretically-motivated feature sets (e.g., latent signal captured by text or speech embeddings), or via correlated proxy features within our feature space; on the latter, we find near-zero correlations between our lexical, acoustic-prosodic, and CSW style feature sets across all corpora and granularities (Appendix~\ref{sec:app_correlations}, Tables~\ref{tab:conv_level_corr}-\ref{tab:turn_level_corr}), suggesting that this alignment gap is not simply an artifact of models exploiting a correlated proxy elsewhere in the same feature space. Second, this design means that we cannot use high feature-importance alignment as straightforward validation, as such alignment could reflect class-definition circularity rather than genuine behavioral correspondence. We therefore treat our findings primarily as evidence of imperfect alignment (the negative result) rather than a strong positive claim about what more obviously aligned model behavior would mean.

\subsection{Cross-lingual entrainment behaviors: effect sizes.}
\label{sec:app_effect_sizes}

Tables~\ref{tab:seame-d}-\ref{tab:masac-lexical-d} report paired Cohen's $d$ for turn- and conversation-level proximity and conversation-level convergence comparisons and lexical entrainment (i.e., proximity) comparisons. Tables~\ref{tab:seame-r} and \ref{tab:masac-r} report aggregate mean and median Pearson $r$ 
for turn-level convergence and synchrony, following standard practice for correlation-based measures. 
These results generally support our main claims in Section~\ref{sec:part1_results} and offer additional evidence beyond what significance testing alone can show. We highlight a few examples below.

First, significant turn-level CSW style proximity effects in SEAME are consistently weak in magnitude ($|d| \leq 0.05$), while the corresponding conversation-level proximity effects are medium-to-large ($0.55 \leq  |d| \leq 1.58$), aligning with greater conversation-level CSW style entrainment within this corpus. Second, MaSaC’s null lexical results for affirmative cues and fillers correspond to effect sizes close to zero ($|d| = 0.01$ and $|d| = 0.02$, respectively), supporting the interpretation that these reflect real absences of entrainment. Third, acoustic-prosodic entrainment effect sizes are uniformly small across every tested feature in SEAME (turn-level $|d| < 0.04$ throughout), providing quantitative support for the characterization of Mandarin-English acoustic-prosodic entrainment as limited, in addition to which individual features crossed the significance threshold. 

\begin{table*}[h]
\centering
\small
\begin{tabular}{@{}lccc@{}}
\toprule
\textbf{Feature} & \textbf{Turn-level Proximity\ ($d$)} & \textbf{Conversation-level Proximity\ ($d$)} & \textbf{Conversation-level Convergence\ ($d$)} \\
\midrule
\multicolumn{4}{@{}l}{\textit{CSW style}} \\
CSW presence & $-0.018$ & $-1.211$ & --- \\
CSW ratio & $-0.050$ & $-1.579$ & --- \\
CSW strategy (\textit{I}) & $-0.016$ & $-0.673$ & --- \\
CSW strategy (\textit{A}) & $-0.0003$ & $-0.552$ & --- \\
CSW strategy (\textit{O}) & $-0.0017$ & $-0.689$ & --- \\
\midrule
\multicolumn{4}{@{}l}{\textit{Acoustic-prosodic}} \\
Min.\ pitch & --- & --- & --- \\
Mean pitch & --- & $0.054$ & --- \\
Max.\ pitch & --- & --- & --- \\
SD pitch & --- & $0.019$ & --- \\
Min.\ intensity & --- & $-0.049$ & --- \\
Mean intensity & $-0.037$ & --- & --- \\
Max.\ intensity & $-0.030$ & --- & --- \\
SD intensity & $-0.0099$ & --- & --- \\
Jitter & --- & $-0.048$ & --- \\
Shimmer & --- & $0.028$ & --- \\
HNR & $-0.021$ & --- & --- \\
Speaking rate & --- & --- & --- \\
\bottomrule
\end{tabular}
\caption{Cohen's $d$ for turn- and conversation-level proximity and conversation-level convergence comparisons in SEAME. ``---'' indicates a feature not identified as significant in Section~\ref{sec:part1_results}'s within-corpus analysis, for which an effect size was therefore not computed.}
\label{tab:seame-d}
\end{table*}

\begin{table}[h]
\centering
\small
\begin{tabular}{@{}lc@{}}
\toprule
\textbf{Feature} & \textbf{Lexical Entrainment\ ($d$)} \\
\midrule
Top-100 words & $1.843$ \\
Top-25 words & $1.795$ \\
Top-25 (conv.-level) & $1.460$ \\
Affirmative cues & $1.679$ \\
Fillers & $0.510$ \\
\bottomrule
\end{tabular}
\caption{Cohen's $d$ for lexical entrainment comparisons in SEAME.}
\label{tab:seame-lexical-d}
\end{table}

\begin{table*}[h]
\centering
\small
\begin{tabular}{@{}lccc@{}}
\toprule
\textbf{Feature} & \textbf{Turn-level Proximity\ ($d$)} & \textbf{Conversation-level Proximity\ ($d$)} & \textbf{Conversation-level Convergence\ ($d$)} \\
\midrule
\multicolumn{4}{@{}l}{\textit{CSW style}} \\
CSW presence & --- & --- & --- \\
CSW ratio & --- & --- & --- \\
CSW strategy (\textit{I}) & --- & --- & --- \\
CSW strategy (\textit{A}) & --- & --- & --- \\
CSW strategy (\textit{O}) & --- & --- & --- \\
\midrule
\multicolumn{4}{@{}l}{\textit{Acoustic-prosodic}} \\
Min.\ pitch & --- & --- & --- \\
Mean pitch & $-0.060$ & $0.026$ & --- \\
Max.\ pitch & --- & --- & --- \\
SD pitch & $-0.027$ & $0.214$ & --- \\
Min.\ intensity & $0.027$ & --- & --- \\
Mean intensity & --- & --- & $-0.155$ \\
Max.\ intensity & --- & --- & $-0.165$ \\
SD intensity & $0.030$ & --- & --- \\
Jitter & $-0.011$ & --- & --- \\
Shimmer & $0.023$ & $0.228$ & --- \\
HNR & $0.029$ & $0.157$ & --- \\
Speaking rate & $0.011$ & $0.148$ & $0.132$ \\
\bottomrule
\end{tabular}
\caption{Cohen's $d$ for turn- and conversation-level proximity and conversation-level convergence comparisons in MaSaC. ``---'' indicates a feature not identified as significant in Section~\ref{sec:part1_results}'s within-corpus analysis, for which an effect size was therefore not computed.}
\label{tab:masac-d}
\end{table*}

\begin{table}[h]
\centering
\small
\begin{tabular}{@{}lc@{}}
\toprule
\textbf{Feature} & \textbf{Lexical Entrainment\ ($d$)} \\
\midrule
Top-100 words & $0.374$ \\
Top-25 words & $0.291$ \\
Top-25 (conv.-level) & $-1.461$ \\
Affirmative cues & $-0.009$ \\
Fillers & $0.015$ \\
\bottomrule
\end{tabular}
\caption{Cohen's $d$ for lexical entrainment comparisons in MaSaC.}
\label{tab:masac-lexical-d}
\end{table}

\begin{table*}[h]
\centering
\small
\begin{tabular}{@{}lcccc@{}}
\toprule
\textbf{Feature family} & \multicolumn{2}{c}{\textbf{Turn-level Convergence}} & \multicolumn{2}{c}{\textbf{Turn-level Synchrony}} \\
 & Mean $r$ & Median $r$ & Mean $r$ & Median $r$ \\
\midrule
CSW style & $-0.015$ & $-0.017$ & $0.031$ & $0.027$ \\
Acoustic-prosodic & $-0.009$ & $-0.017$ & $0.012$ & $0.009$ \\
\bottomrule
\end{tabular}
\caption{Mean and median Pearson's $r$ for turn-level convergence and synchrony in SEAME, aggregated by feature family.}
\label{tab:seame-r}
\end{table*}

\begin{table*}[h]
\centering
\small
\begin{tabular}{@{}lcccc@{}}
\toprule
\textbf{Feature family} & \multicolumn{2}{c}{\textbf{Turn-level Convergence}} & \multicolumn{2}{c}{\textbf{Turn-level Synchrony}} \\
 & Mean $r$ & Median $r$ & Mean $r$ & Median $r$ \\
\midrule
CSW style & $0.070$ & $0.080$ & $-0.259$ & $-0.059$ \\
Acoustic-prosodic & $0.017$ & $0.023$ & $0.006$ & $-0.001$ \\
\bottomrule
\end{tabular}
\caption{Mean and median Pearson's $r$ for turn-level convergence and synchrony in MaSaC, aggregated by feature family.}
\label{tab:masac-r}
\end{table*}

\subsection{Cross-lingual entrainment behaviors: Bonferroni-corrected significance results.}
\label{sec:app_bonferroni}

Table~\ref{tab:cross-corpus-corrected} summarizes cross-corpus Bonferroni-corrected pairwise Fisher's exact comparisons; Table~\ref{tab:within-corpus-corrected} summarizes within-corpus Bonferroni-corrected significance for the same measures. See Section~\ref{sec:part1_results} for relevant interpretation.

\begin{table*}[h]
\centering
\small
\begin{tabular}{@{}ccc p{10cm}@{}}
\toprule
\textbf{Feature family} & \textbf{Surviving} & \textbf{Total} & \textbf{Notable corrected results} \\
\midrule
Lexical & 1 & 18 & Only one contrast (top-100 corpus-level words, SEAME vs. MaSaC) survives correction, supporting the claim that lexical entrainment generalizes robustly across language pairs. \\
Acoustic-prosodic & 11 & 178 & The majority of surviving contrasts involve SEAME diverging from both BM and MaSaC, consistent with the characterization of Mandarin-English acoustic-prosodic patterns as the most distinct of the three pairs. \\
CSW style & 12 & 60 & Surviving contrasts are concentrated in the features we highlight in Section~\ref{sec:part1_results}: MaSaC vs.\ BM and MaSaC vs.\ SEAME differ significantly on conversation-level alternational CSW proximity ($p_{corrected}<0.001$, both) and turn-level insertional CSW synchrony ($p_{corrected}<0.001$, both), while BM and SEAME remain statistically indistinguishable from each other on the same features. This is consistent with the characterization of MaSaC's CSW style entrainment as markedly different from the other two corpora.\\
\bottomrule
\end{tabular}
\caption{Cross-corpus Bonferroni-corrected pairwise comparisons by feature family.}
\label{tab:cross-corpus-corrected}
\end{table*}

\begin{table*}[h]
\centering
\small
\begin{tabular}{@{}cccc p{9cm}@{}}
\toprule
\textbf{Corpus} & \textbf{Feature family} & \textbf{Surviving} & \textbf{Total} &  \textbf{Notable corrected results} \\
\midrule
SEAME & Lexical & 6 & 6 & All five word classes and overall language use remain significant ($p_{corrected}<0.001$, each). \\
\addlinespace
MaSaC & Lexical & 4 & 6 & Four of six features remain significant: corpus-level top-100 and top-25 words, conversation-level top-25 words, and overall language use ($p_{corrected}<0.01$, each); affirmative cues and fillers do not survive correction ($p_{corrected}=1.0$, both). This indicates that the Hindi-English lexical entrainment asymmetry reported in Section~\ref{sec:part1_results} is robust to multiple-comparison correction rather than an artifact of uncorrected significance testing. \\
\midrule
SEAME & Acoustic-prosodic & 17 & 60 & Few turn-level proximity contrasts survive correction (3 of 12 features: mean and maximum intensity and HNR), consistent with the reported pattern of inconsistent proximity in Mandarin-English. Conversation-level proximity and convergence largely fail to survive correction, supporting the claim that Mandarin-English mirrors the absence of conversation-level acoustic-prosodic entrainment previously reported for Spanish-English.\\
\addlinespace
MaSaC & Acoustic-prosodic & 26 & 60 & Turn-level proximity contrasts remain overwhelmingly significant (10 of 12 features: all except minimum pitch and maximum intensity; $p_{corrected}<0.05$), supporting the claim of substantially stronger acoustic-prosodic entrainment in Hindi-English than in Mandarin-English. All turn-level convergence contrasts survive correction (12 of 12 features; $p_{corrected}<0.05$, each). Conversation-level proximity and convergence largely fail to survive correction, supporting the claim that Hindi-English mirrors the absence of conversation-level acoustic-prosodic entrainment previously reported for Spanish-English.
\\
\midrule
SEAME & CSW style & 9 & 20 & Significant contrasts that survive correction are concentrated in turn- and conversation-level proximity, supporting the findings reported in Section~\ref{sec:part1_results}.\\
\addlinespace
MaSaC & CSW style & 4 & 20 & Significant contrasts that survive correction are concentrated exclusively in turn-level convergence across CSW presence, ratio, and insertional strategy ($p_{corrected}<0.001$, each); no other measure or entrainment level survives correction. This supports the claim that Hindi-English CSW style entrainment in MaSaC is distinctively convergence-driven, in contrast to the proximity- and synchrony-dominant patterns observed observe for Mandarin-English (and those previously reported for Spanish-English). 
\\
\bottomrule
\end{tabular}
\caption{Within-corpus Bonferroni-corrected significance by feature family.}
\label{tab:within-corpus-corrected}
\end{table*}

\subsection{Cross-lingual entrainment behaviors: additional results.}
\label{sec:app_acoustic-prosodic-seame}

\paragraph{Entrainment over lexical aspects of code-switched speech.}In the SEAME corpus, when excluding OOV words, 88\% of conversations exhibit within-conversation entrainment ($p < 0.05$); 82\% involve both partners entraining to each other, while in the remaining 6\% only one speaker entrains to their partner. 

When excluding OOV words in the MaSaC corpus, 72\% of conversations exhibit entrainment ($p < 0.05$); 58\% involve both interlocutors entraining to each other, while the remaining 14\% involve only unidirectional entrainment.

See Figures~\ref{fig:proximity-turn-seame} through \ref{fig:synchrony-turn-seame} for visualizations of turn-level proximity, convergence, and synchrony in SEAME. 

\begin{figure}[H]
\includegraphics[width=\columnwidth]{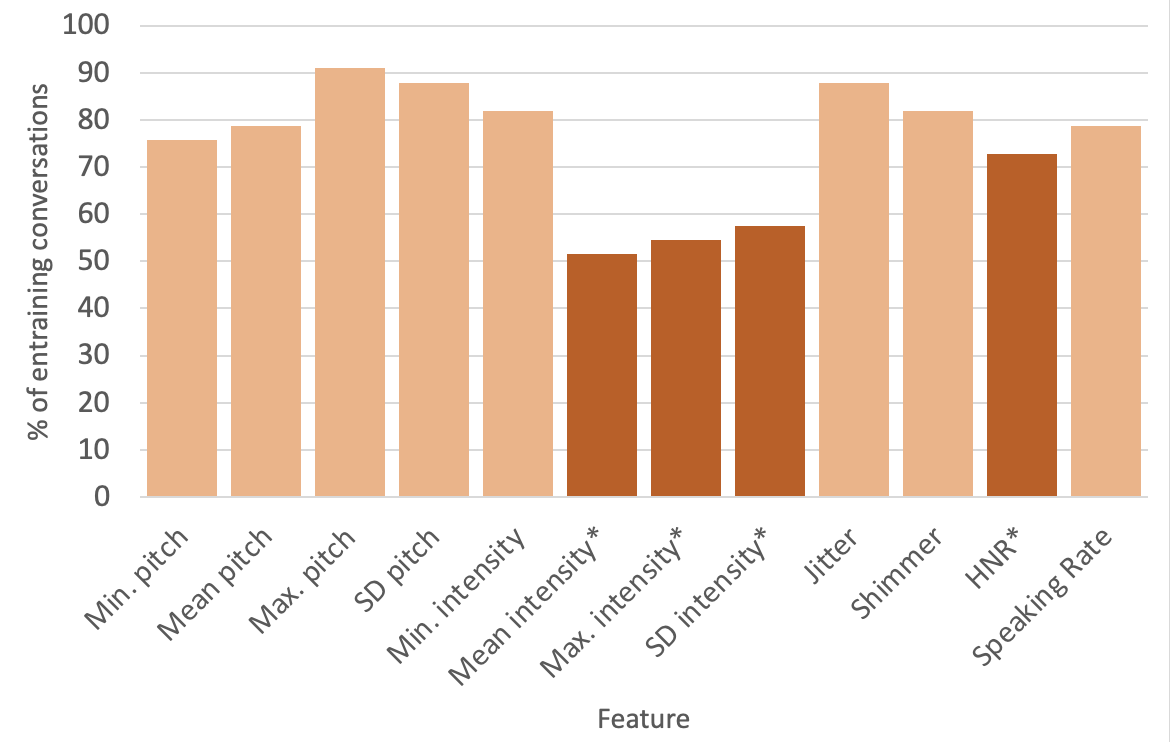}
    \caption{Proximity at the turn-level in SEAME. * and darker bars indicate significant features.}  
    \label{fig:proximity-turn-seame}
\end{figure}

\begin{figure}[H]
\includegraphics[width=\columnwidth]{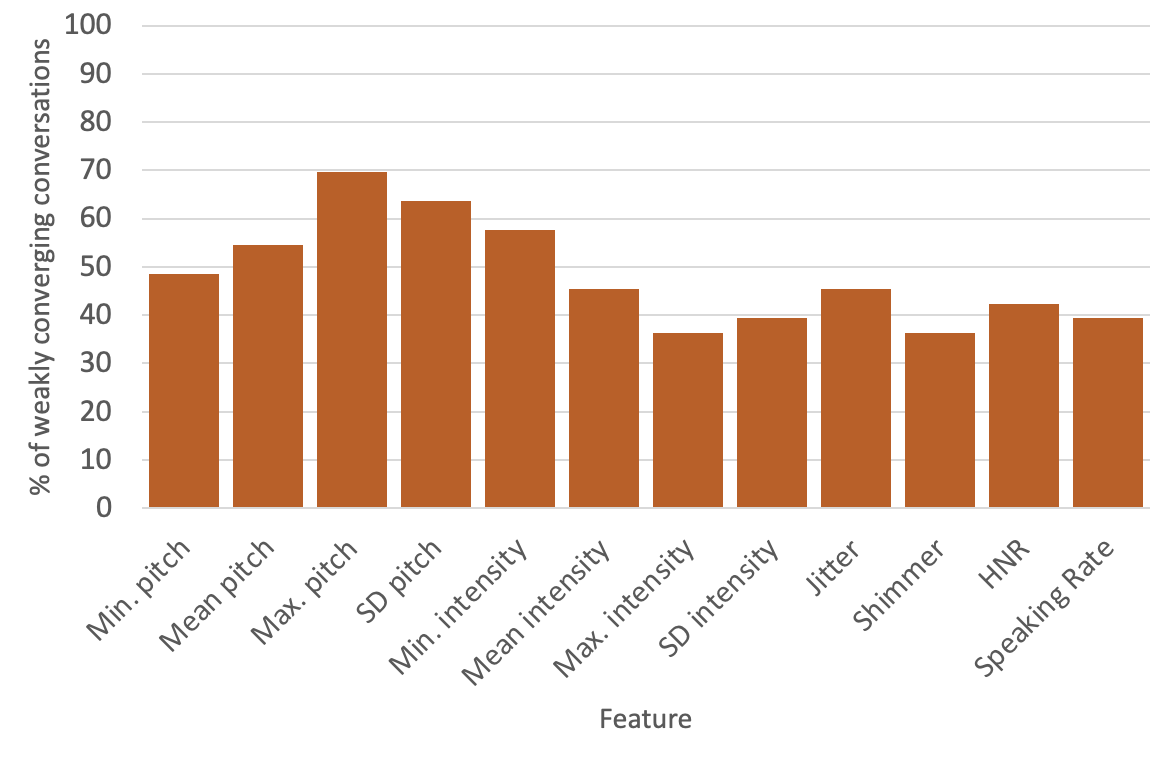}
    \caption{Convergence at the turn-level in SEAME. All conversations converge \textit{weakly}. All non-converging conversations diverge \textit{weakly}. All acoustic-prosodic features are significant.}  
    \label{fig:convergence-turn-seame}
\end{figure}

\begin{figure}[H]
\includegraphics[width=\columnwidth]{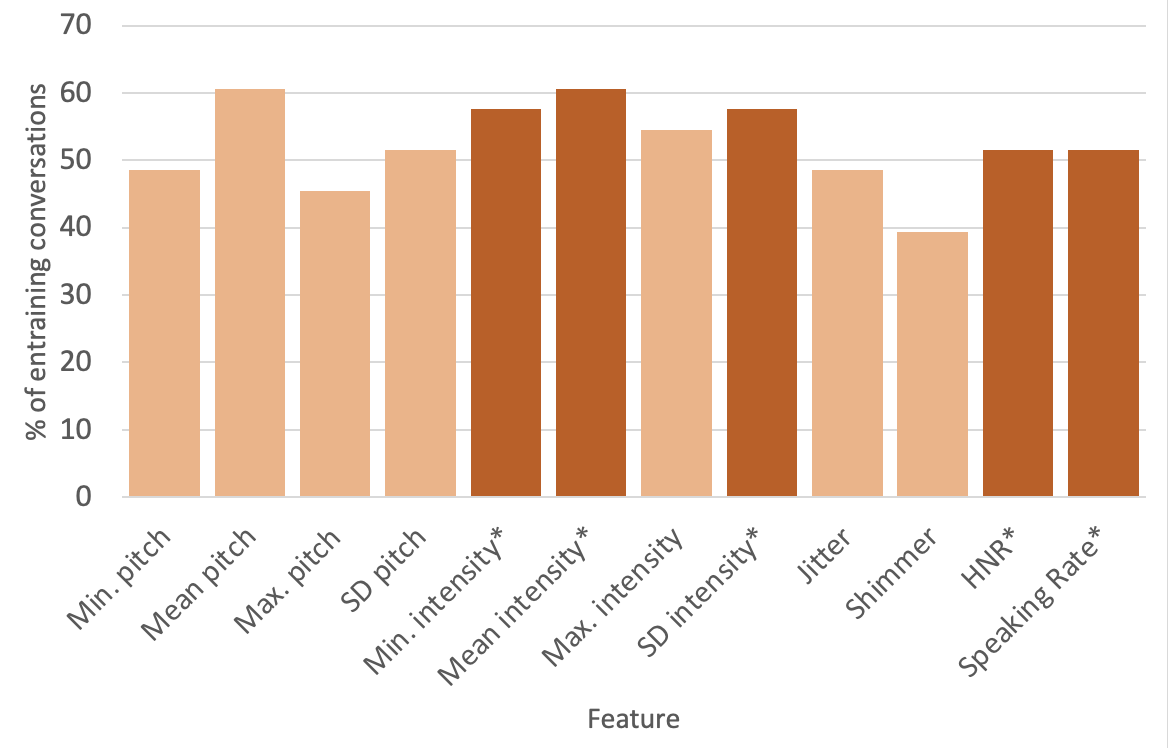}
    \caption{Synchrony at the turn-level in SEAME. All conversations are \textit{weakly} synchronous. All acoustic-prosodic features are significant. All non-synchronous conversations are \textit{weakly} asynchronous.}  
    \label{fig:synchrony-turn-seame}
\end{figure}

\subsection{Cross-lingual entrainment behaviors: comparison to prior monolingual and CSW entrainment findings.}
\label{sec:app_corpus_comparisons}

While exact quantitative comparisons are limited by differences in corpora, modality, and entrainment operationalization across studies, we offer several qualitative points of comparison to prior work below.

\citet{bawa-etal-2018-accommodation} study Hindi-English CSW accommodation in a Hindi movie corpus using a distinct entrainment metric from our own. Their finding that a high base rate of CSW corresponds to reduced accommodation is consistent with our hypothesis (Section~\ref{sec:part1_results}) that MaSaC's higher M- and I-index values may create a ceiling on stylistic entrainment, since there is less room for one speaker's style to influence the other's when both already code-switch prevalently.

\citet{levitan-etal-2015-acoustic} study monolingual acoustic-prosodic entrainment in Spanish, English, and Mandarin. Their finding of significant monolingual English turn-level proximity on mean and maximum intensity aligns directionally with our finding that SEAME's limited turn-level proximity is nonetheless significant specifically on intensity features (Figure~\ref{fig:proximity-turn-seame}), suggesting that the English side of the language pair may drive this signal. Their slight yet largely negative Mandarin proximity and strongly negative Mandarin pitch synchrony are consistent with our characterization of Mandarin-English as our weakest acoustic-prosodic pair, lending external support to our explanation involving Mandarin's distinct tonal character.

\citet{xia14_speechprosody} report that monolingual Mandarin and English each individually show strong conversation-level proximity on intensity, pitch maxima, and speaking rate, with English additionally showing conversation-level convergence. Our SEAME findings show the opposite pattern: an absence of conversation-level acoustic-prosodic proximity and convergence, mirroring Spanish-English. This divergence suggests that the CSW context itself, rather than either language alone, may suppress conversation-level entrainment that is otherwise present monolingually in both languages.

\citet{parekh-etal-2020-understanding} study lexical and CSW entrainment in written human-machine Hindi-English dialogue. Their finding of strong lexical entrainment in both English and Hindi, using the same method as \citet{nenkova-etal-2008-high} that we also adopt, is broadly consistent with our MaSaC lexical findings. We also note a partial parallel between their reported entrainment ranking (CSW presence, then insertional CSW, then language choice, then alternational CSW) and our finding that MaSaC's turn-level convergence is strongest for CSW presence and insertional strategy; however, their study concerns accommodation to a machine interlocutor in text rather than human-human speech, likely reflecting a different underlying social dynamic.

\subsection{Classifier models' entrainment behavior: models' feature importance values.}
\label{sec:app_feature_importances}

Figure~\ref{fig:bm-feature-importances-full} shows the full set of feature importance values across the best-performing BM models at the conversation-level, including many near-zero values. Figures~\ref{fig:seame-feature-importances} and \ref{fig:seame-feature-importances-turnlevel} visualize conversation-level and turn-level features importance values, respectively, for the SEAME corpus. Figures~\ref{fig:masac-feature-importances} and \ref{fig:masac-feature-importances-turnlevel} visualize the same in MaSaC. 

With respect to the degree of importance assigned to CSW style features in the best-performing turn-level speech-embedding models for MaSaC (Figure~\ref{fig:masac-feature-importances-turnlevel}), this could partly reflect very weak turn-level cross-feature-set correlations  (Table~\ref{tab:turn_level_corr}; Appendix~\ref{sec:app_correlations}). Although the mean correlation between CSW metrics and acoustic-prosodic features is near zero ($r=0.015$), 35.4\% of feature pairs reach statistical significance; this is likely an artifact of the large number of turn pairs inflating test sensitivity rather than evidence of a meaningful relationship. However, even such weak statistical regularities can be sufficient for models to exploit CSW features as incidental predictors of acoustic-prosodic entrainment labels. We highlight that this is importantly different from our interest in models learning a theoretically-motivated entrainment signal.

\begin{figure*}
\centering
\includegraphics[width=\linewidth]{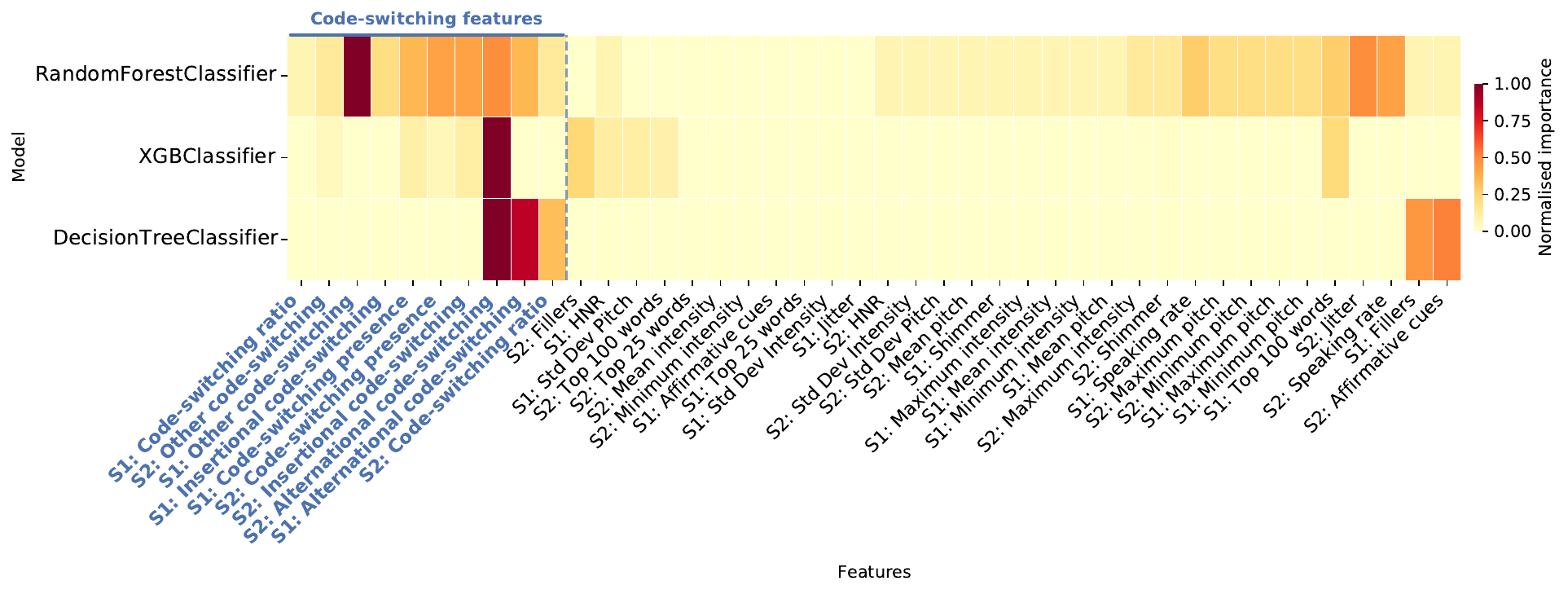}
    \caption{Feature importance values across best-performing BM models at the conversation-level. We normalize values per model using row-wise min.-max. so that models with different importance scales are comparable.
    }  
    \label{fig:bm-feature-importances-full}
    \vspace{-2mm}
\end{figure*}

\begin{figure*}[!h]
\centering
\includegraphics[width=\linewidth]{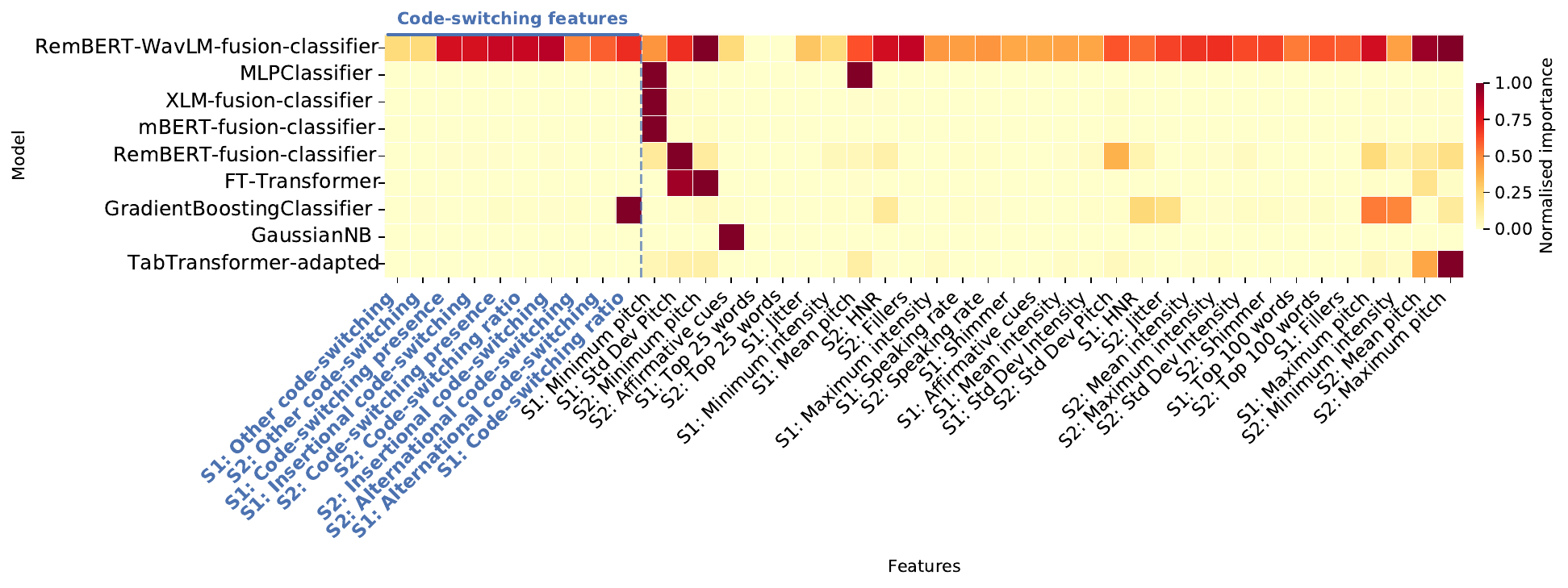}
    \caption{Feature importance values across best-performing SEAME models at the conversation-level. We normalize values per model using row-wise min.-max. so that models with different importance scales are comparable. All models except \textit{FT-Transformer} achieve near-perfect test-time accuracy; \textit{FT-Transformer achieves 75\%} test-time accuracy. Individual model hyperparameter settings are as follows: \textit{MLPClassifier} has activation: tanh, alpha: 1e-05, hidden layer sizes: 250, solver: lbfgs; \textit{GradientBoostingClassifier} has criterion: friedman mse, lr: 0.01, loss: exponential, n estimators: 50; \textit{TabTransformer-adapted} has lr: 0.001; \textit{XLM-fusion} has lr: 0.001; \textit{mBERT-fusion} has lr: 0.001; \textit{RemBERT-fusion} has lr: 0.001; \textit{RemBERT + WavLM fusion} has lr: 0.01; \textit{FT-Transformer} has lr: 0.001.}  
    \label{fig:seame-feature-importances}
\end{figure*}

\begin{figure*}[!h]
\centering
\includegraphics[width=\linewidth]{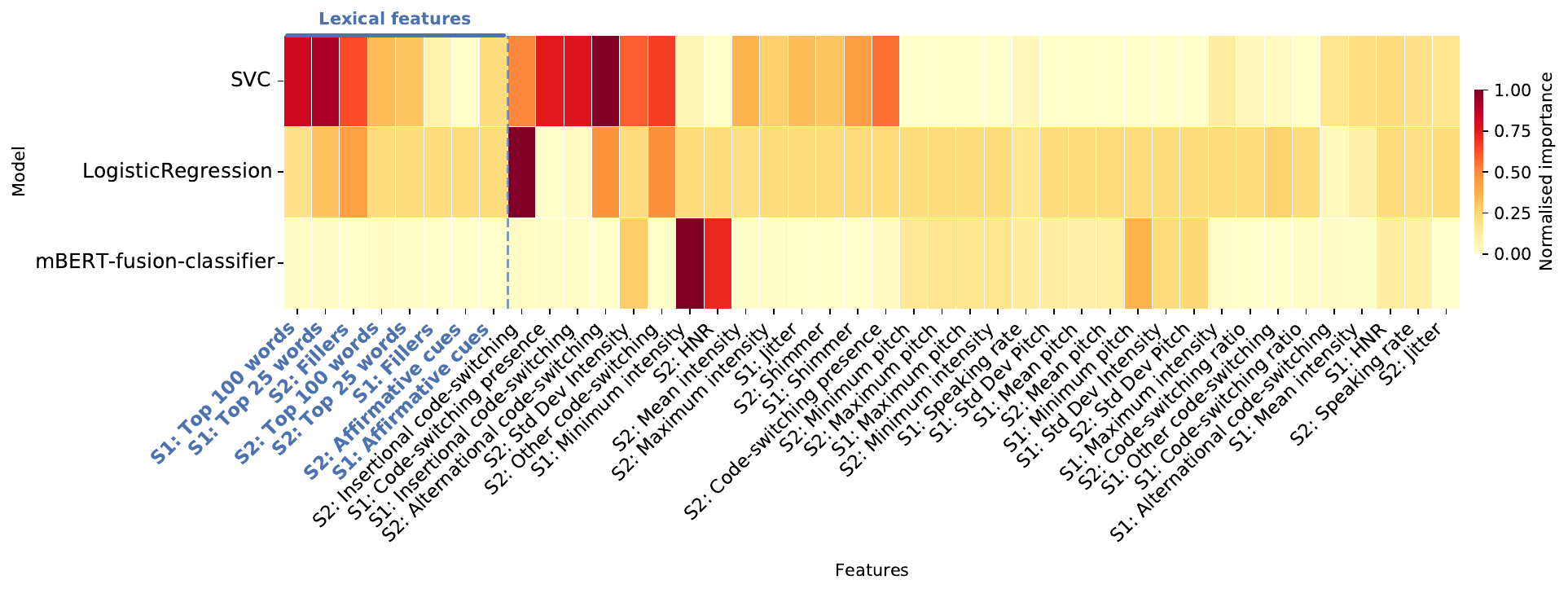}
    \caption{Feature importance values across best-performing MaSaC models at the conversation-level. We normalize values per model using row-wise min.-max. so that models with different importance scales are comparable. \textit{LogisticRegression} with hyperparameter C: 1 achieves 71\% test-time accuracy, \textit{mBERT-fusion} with hyperparameter lr: 0.001 achieves 68\%, and \textit{SVC} with hyperparameter C: 1.75 achieves 65\%.}  
    \label{fig:masac-feature-importances}
\end{figure*}

\begin{figure*}[!h]
\centering
\includegraphics[width=\linewidth]{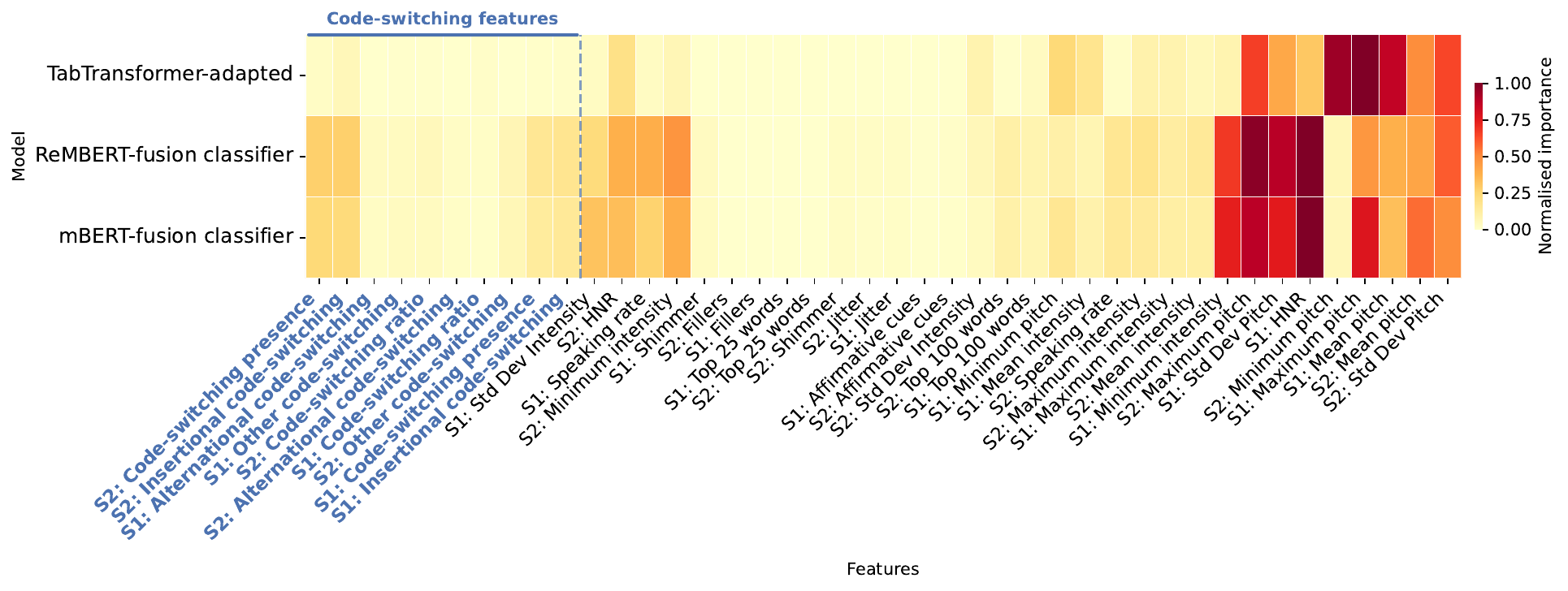}
    \caption{Feature importance values across best-performing SEAME models at the turn-level. We normalize values per model using row-wise min.-max. so that models with different importance scales are comparable. \textit{mBERT-fusion} with hyperparameter lr: 1e-4 achieves 70\% test-time accuracy, \textit{RemBERT-fusion} with hyperparameter lr: 1e-4 achieves 67\%, and \textit{TabTransformer-adapted} with hyperparameter lr: 1e-3  achieves 65\%.}  
    \label{fig:seame-feature-importances-turnlevel}
\end{figure*}

\begin{figure*}[!h]
\centering
\includegraphics[width=\linewidth]{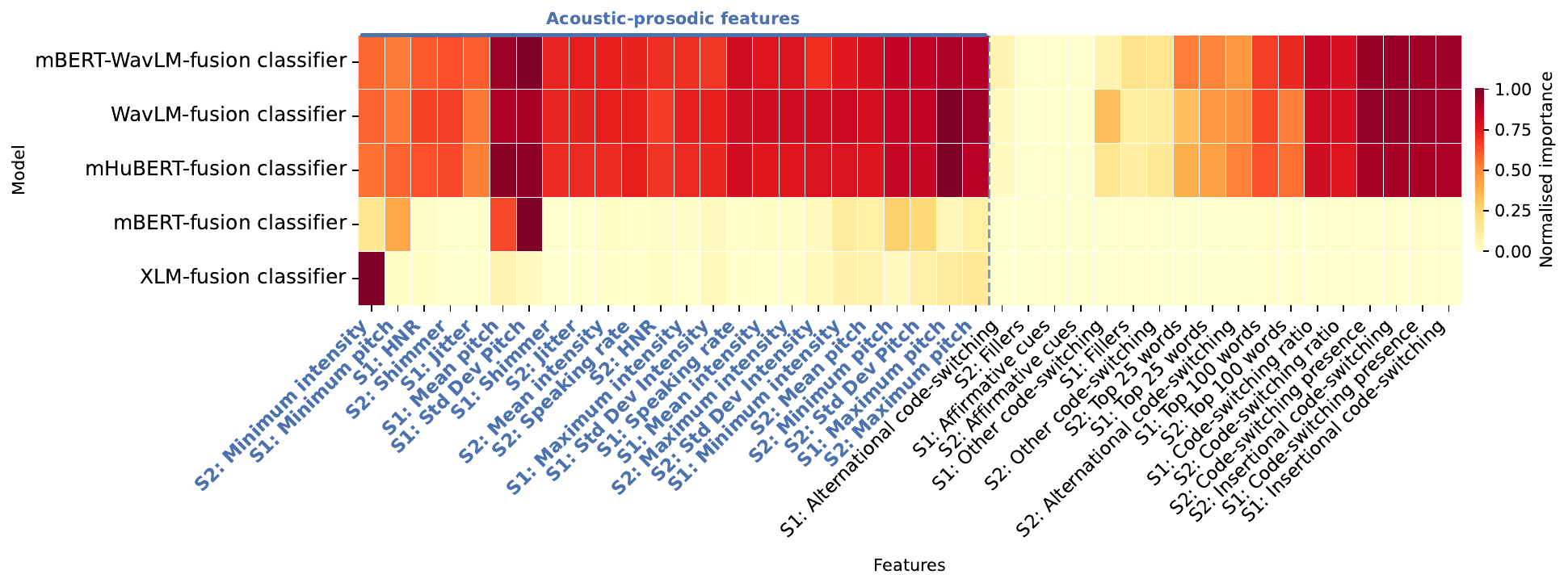}
    \caption{Feature importance values across best-performing MaSaC models at the turn-level. We normalize values per model using row-wise min.-max. so that models with different importance scales are comparable. \textit{mBERT-fusion} with hyperparameter lr: 2e-5 achieves 70\% test-time accuracy, \textit{XLM-fusion} with hyperparameter lr: 5e-5 achieves 69\%, \textit{WavLM-fusion} with hyperparameter lr: 1e-3 achieves 74\%, \textit{mHuBERT-fusion} with hyperparameter lr: 1e-3 achieves 70\%, and \textit{mBERT+WavLM-fusion} with hyperparameter lr: 1e-3 achieves 77\%.}  
    \label{fig:masac-feature-importances-turnlevel}
\end{figure*}

\subsection{Classifier models' entrainment behavior: feature-level ablations.}
\label{sec:app_ablations}

Figures~\ref{fig:seame-ablations} and \ref{fig:seame-ablations-turnlevel} visualize feature ablations on the SEAME corpus at the conversation-level and the turn-level, respectively. Figures~\ref{fig:masac-ablations} and \ref{fig:masac-ablations-turnlevel} visualize the same on the MaSaC corpus. 

\begin{figure*}[!h]
\centering
\includegraphics[width=\linewidth]{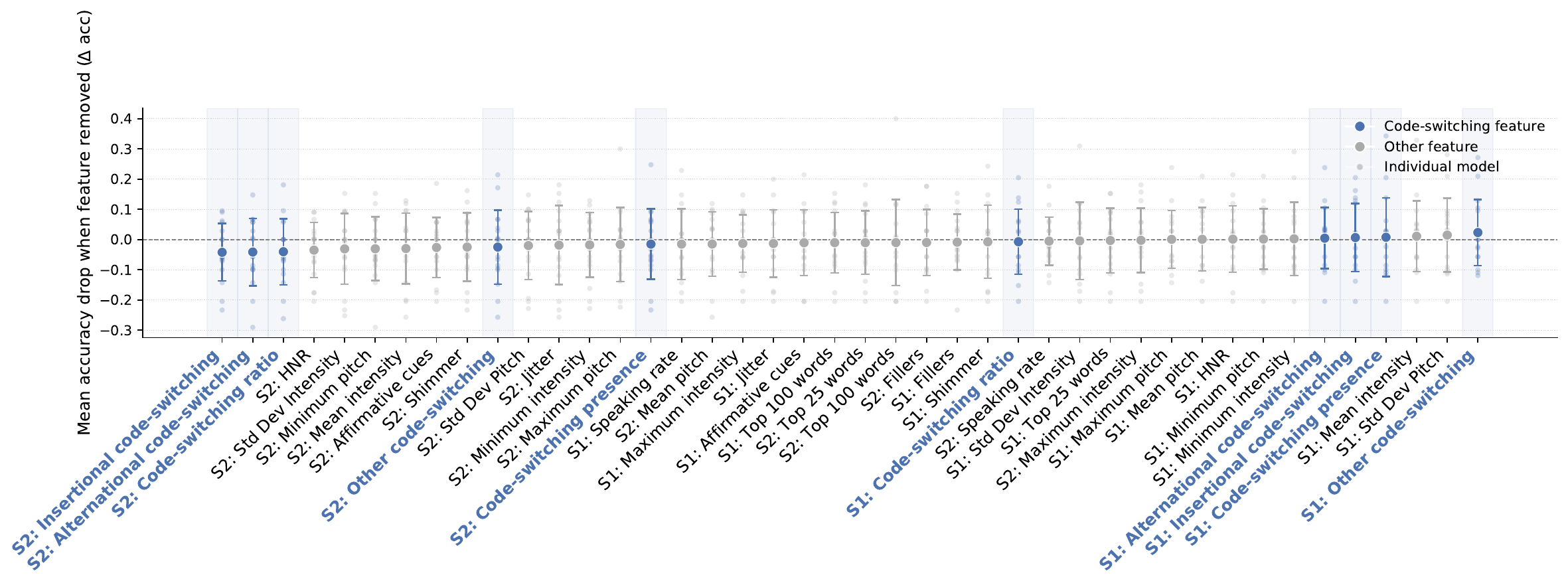}
    \caption{Aggregated feature ablation results for SEAME at the conversation-level. Features are ordered left to right by increasing impact; 
    $\Delta_{acc.}>0$ means accuracy \textit{drops} when the feature is removed, i.e., the feature is useful. }
    \label{fig:seame-ablations}
\end{figure*}

\begin{figure*}[!h]
\centering
\includegraphics[width=\linewidth]{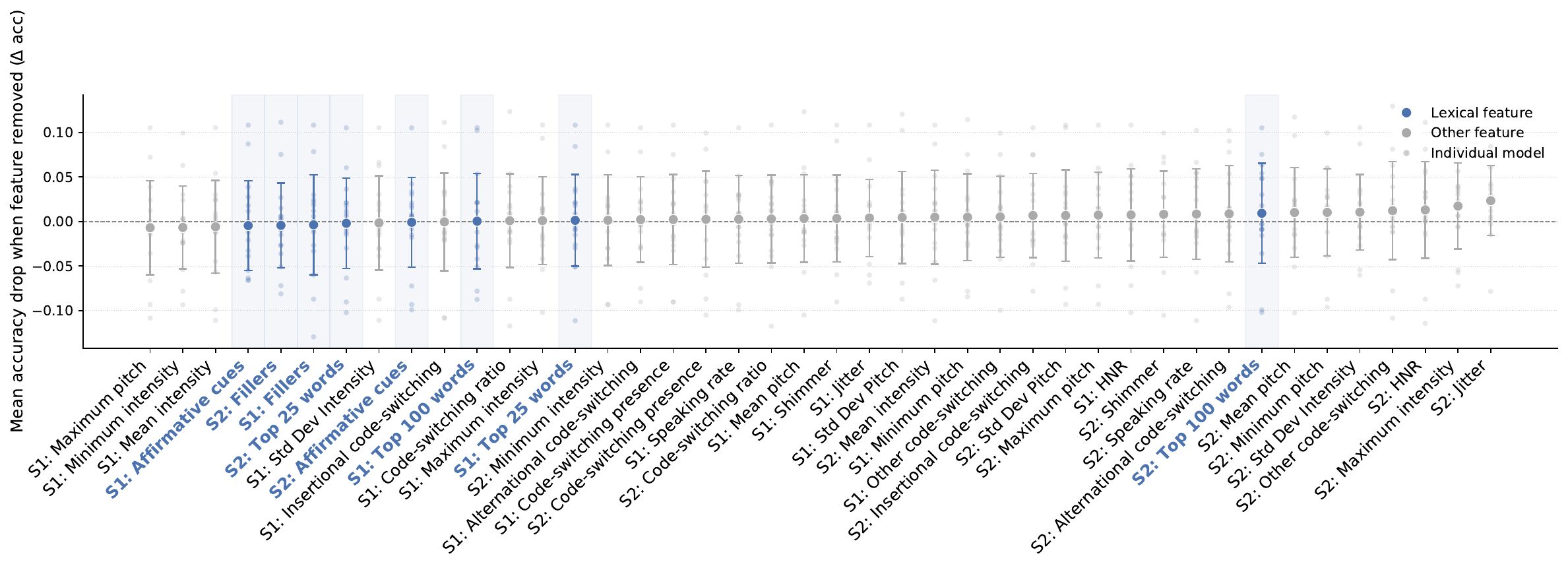}
    \caption{Aggregated feature ablation results for MaSaC at the conversation-level. Features are ordered left to right by increasing impact; 
    $\Delta_{acc.}>0$ means accuracy \textit{drops} when the feature is removed, i.e., the feature is useful. }
    \label{fig:masac-ablations}
\end{figure*}

\begin{figure*}[!h]
\centering
\includegraphics[width=\linewidth]{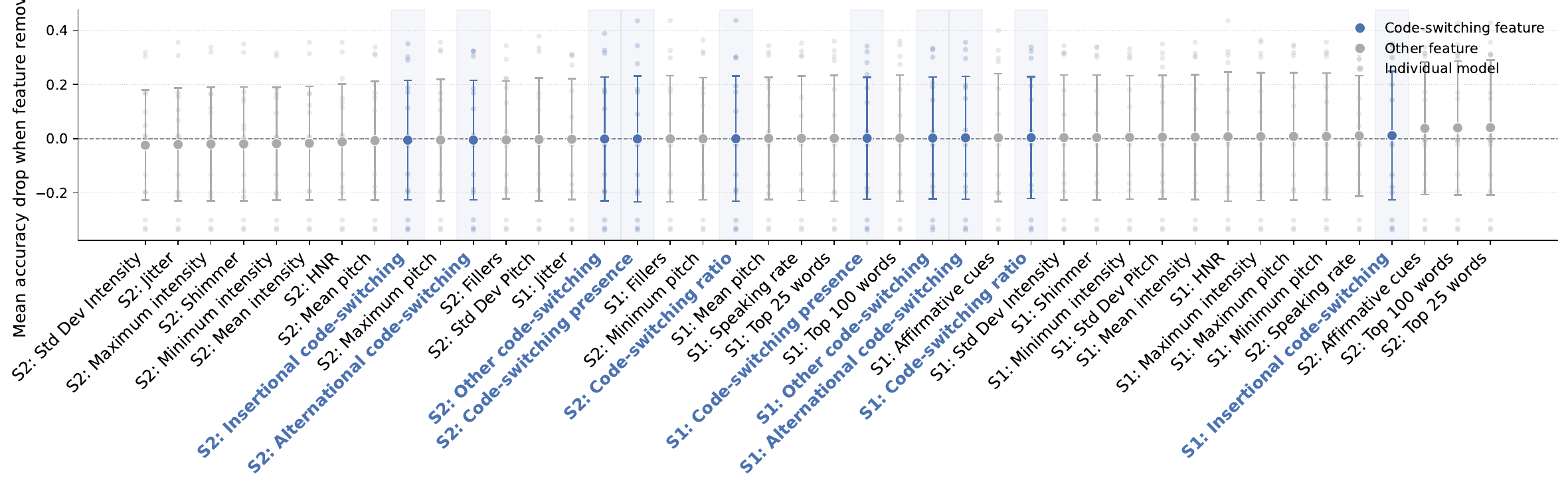}
    \caption{Aggregated feature ablation results for SEAME at the turn-level. Features are ordered left to right by increasing impact; 
    $\Delta_{acc.}>0$ means accuracy \textit{drops} when the feature is removed, i.e., the feature is useful. }
    \label{fig:seame-ablations-turnlevel}
\end{figure*}

\begin{figure*}[!h]
\centering
\includegraphics[width=\linewidth]{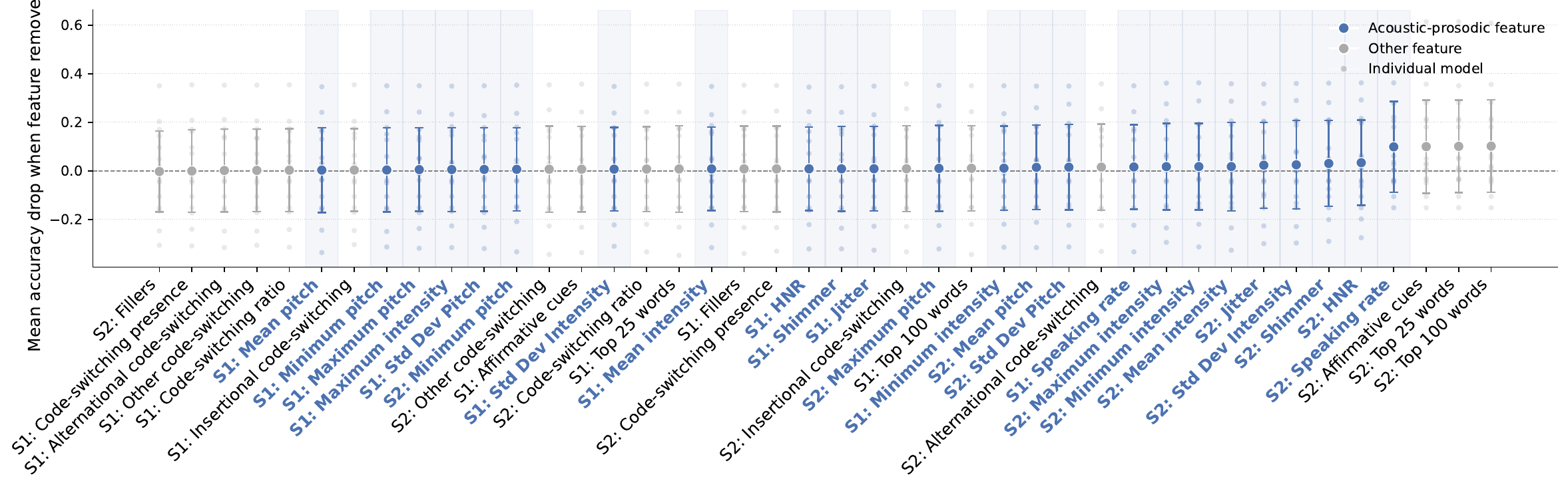}
    \caption{Aggregated feature ablation results for MaSaC at the turn-level. Features are ordered left to right by increasing impact; 
    $\Delta_{acc.}>0$ means accuracy \textit{drops} when the feature is removed, i.e., the feature is useful. }
    \label{fig:masac-ablations-turnlevel}
\end{figure*}

\begin{figure*}[!h]
\centering
\includegraphics[width=\linewidth]{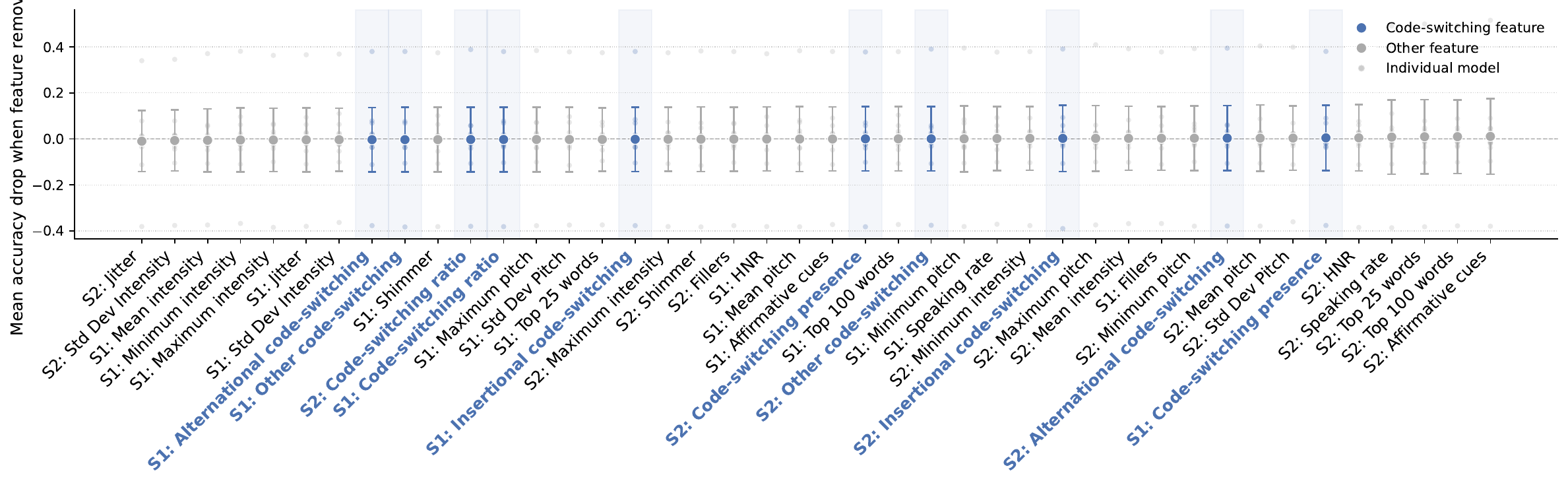}
    \caption{Aggregated feature ablation results for BM at the turn-level. Features are ordered left to right by increasing impact; 
    $\Delta_{acc.}>0$ means accuracy \textit{drops} when the feature is removed, i.e., the feature is useful. }
    \label{fig:bm-ablations-turnlevel}
\end{figure*}

\begin{table}[!htbp]
\footnotesize
\centering
  \begin{tabular}{ cccccc }
    \toprule
    \textbf{Corpus} &\textbf{FS 1}  & \textbf{FS 2} & \textbf{Mean $r$} & \textbf{SD $r$} & \textbf{\% sig.}\\
    \midrule
    BM & CSW & CSW & 0.547 & 0.259 & 53.3 \\
    BM & L & L & -0.117 & 0.329 & 46.7 \\
    BM & AP & AP & 0.369 & 0.310 & 51.8 \\
    BM & CSW & L & 0.003 & 0.138 & 1.67 \\
    BM & CSW & AP & -0.083 & 0.155 & 10.4 \\
    BM & L & AP & -0.023 & 0.202 & 9.72 \\
    SEAME & CSW & CSW & 0.573 & 0.434 & 46.7 \\
    SEAME & L & L & -0.063 & 0.310 & 20.0 \\
    SEAME & AP & AP & 0.209 & 0.326 & 23.9 \\
    SEAME & CSW & L & 0.072 & 0.350 & 31.7 \\
    SEAME & CSW & AP & -0.021 & 0.231 & 13.8 \\
    SEAME & L & AP & -0.052 & 0.205 & 16.0 \\
    MaSaC & CSW & CSW & 0.197 & 0.297 & 24.4 \\
    MaSaC & L & L & -0.024 & 0.165 & 40.0 \\
    MaSaC & AP & AP & 0.057 & 0.294 & 63.8 \\
    MaSaC & CSW & L & 0.050 & 0.133 & 22.0 \\
    MaSaC & CSW & AP & 0.016 & 0.073 & 11.3 \\
    MaSaC & L & AP & 0.002 & 0.086 & 15.8 \\
    \bottomrule
  \end{tabular}
  \caption{Summary of conversation-level feature set correlations. FS = feature set; CSW = code-switching style aspects; L = lexical features; AP = acoustic-prosodic features.}
    \label{tab:conv_level_corr}
\end{table}

\begin{table}[!htbp]
\footnotesize
\centering
  \begin{tabular}{ cccccc }
    \toprule
    \textbf{Corpus} &\textbf{FS 1}  & \textbf{FS 2} & \textbf{Mean $r$} & \textbf{SD $r$} & \textbf{\% sig.}\\
    \midrule
    BM & CSW & CSW & 0.191 & 0.251 & 42.2 \\
    BM & L & L & 0.185 & 0.251 & 57.1 \\
    BM & AP & AP & 0.249 & 0.236 & 86.6 \\
    BM & CSW & L & 0.026 & 0.049 & 46.3 \\
    BM & CSW & AP & 0.001 & 0.029 & 18.8 \\
    BM & L & AP & 0.019 & 0.047 & 32.8 \\
    SEAME & CSW & CSW & 0.159 & 0.269 & 44.4 \\
    SEAME & L & L & -0.030 & 0.058 & 40.0 \\
    SEAME & AP & AP & 0.153 & 0.231 & 84.8  \\
    SEAME & CSW & L & 0.024 & 0.105 & 53.3  \\
    SEAME & CSW & AP & 0.017 & 0.072 & 68.3 \\
    SEAME & L & AP & 0.012 & 0.084 & 49.3 \\
    MaSaC & CSW & CSW & 0.182 & 0.295 & 46.7 \\
    MaSaC & L & L & 0.162 & 0.242 & 33.3 \\
    MaSaC & AP & AP & 0.056 & 0.263 & 70.7 \\
    MaSaC & CSW & L & -0.019 & 0.052 & 24.3 \\
    MaSaC & CSW & AP & 0.015 & 0.083 & 35.4 \\
    MaSaC & L & AP & 0.004 & 0.052 & 20.2 \\
    \bottomrule
  \end{tabular}
  \caption{Summary of turn-level feature set correlations. FS = feature set; CSW = code-switching style aspects; L = lexical features; AP = acoustic-prosodic features.}
    \label{tab:turn_level_corr}
\end{table}

\subsection{Within- and between-feature-set correlations for each corpus.}
\label{sec:app_correlations}

At the \textbf{conversation-level}, cross-feature-set correlations are near zero across all modality pairs and corpora, with mean Fisher-$z$-transformed correlations close to zero and very low proportions of significant relationships where $p<0.05$ (Table~\ref{tab:conv_level_corr}). This suggests that CSW, lexical, and acoustic-prosodic feature spaces capture largely non-overlapping dimensions of variation, supporting their complementarity rather than redundancy. In contrast, within-set correlations are substantially higher, particularly for CSW and acoustic features in BM and SEAME, indicating weak to moderate internal structure but cross-modally distinct feature spaces in those corpora. Conversation-level patterns of cross-feature-set correlations are consistent at the \textbf{turn-level} (Table~\ref{tab:turn_level_corr}), where even within-set correlations are weak.

\subsection{Classifier models' entrainment behavior: excluding BM from turn-level analysis of models' feature importance values.}
\label{sec:app_BM_exclusion}

None of the BM models exceed the 65\% test-time accuracy threshold at the turn level, so we exclude the corpus from turn-level analysis. This poor performance may be due to class imbalance artifacts and/or the relative size of the corpus, especially given BM's lower overall CSW rate (5\%, Table~\ref{tab:corpora}), which could reduce the variance in CSW style features at the turn level and may make the classification boundary harder to learn. We acknowledge that the exclusion of the corpus is an unfortunate limitation, as the absence of turn-level BM results leaves the cross-lingual picture at finer conversational granularity incomplete.

\subsection{Classifier models' entrainment behavior: conversation-level analysis using CSW style features in MaSaC.}
\label{sec:app_noisy_analysis}

Our primary classification tasks for each corpus use the empirically-determined feature set that best discriminates entraining from non-entraining conversations (Footnote~\ref{footnote:5}); we additionally explore classifier behavior when the feature set in use is not naturally suited to this purpose in a given corpus. While CSW style metrics are appropriate for evaluating proximity in BM and SEAME at the conversation-level, these do not yield a naturally balanced classification task for MaSaC, as the corpus lacks significant conversation-level proximity over this feature set (Sections~\ref{sec:method_models} and \ref{sec:part1_results}). Therefore, here, we construct an \textit{artificial} entraining/non-entraining split for MaSaC by taking the median of an aggregate score across CSW presence, ratio, and strategy, and assigning conversations above the median to the positive class. We otherwise follow the same feature construction, speaker-averaging, and grid search procedure described in Section~\ref{sec:method_models} and Appendix~\ref{sec:app_posnegclasses}.

We caution that this analysis differs in an important respect from our primary classification tasks. Since the entraining/non-entraining label used here is constructed directly as a threshold on the average of the same CSW style features that are also included in the classifier's input feature vector, the resulting feature importance and ablation results (Figures~\ref{fig:masac-feature-importances-noisy} and \ref{fig:masac-ablations-noisy}) are expected to concentrate heavily on these features by construction, independent of any genuine model sensitivity to human-relevant entraining behavior. This is distinct from the circularity concern we raise for our primary analyses in Appendix~\ref{sec:app_posnegclasses}: there, the positive class is defined by a statistical significance test over observed human entrainment, a non-trivial target that a classifier is not guaranteed to recover from the raw feature values alone. Here, in contrast, the label is a direct arithmetic function of a subset of the input features themselves, so strong classifier reliance on those features reflects the construction of the task rather than a positive, human-grounded finding.

Consistent with this, Figure~\ref{fig:masac-feature-importances-noisy} shows near-maximal importance concentrated almost entirely on the CSW style features across models, and Figure~\ref{fig:masac-ablations-noisy} shows a similar pattern: three of the five features with the largest ablation impact are CSW style features. Together, these results reflect the label's own construction rather than demonstrating that classifiers attend to human-relevant CSW style features when detecting entrainment -- a direct consequence of MaSaC's lack of significant conversation-level CSW style proximity, noted above. We report this analysis for completeness and to illustrate the effect of feature-label circularity when it is present in an unambiguous form, but we do not interpret these results as evidence of human-grounded model behavior, and we caution against comparing them directly to the findings reported in Section~\ref{sec:part2_results}.

\begin{figure*}[!h]
\centering
\includegraphics[width=\linewidth]{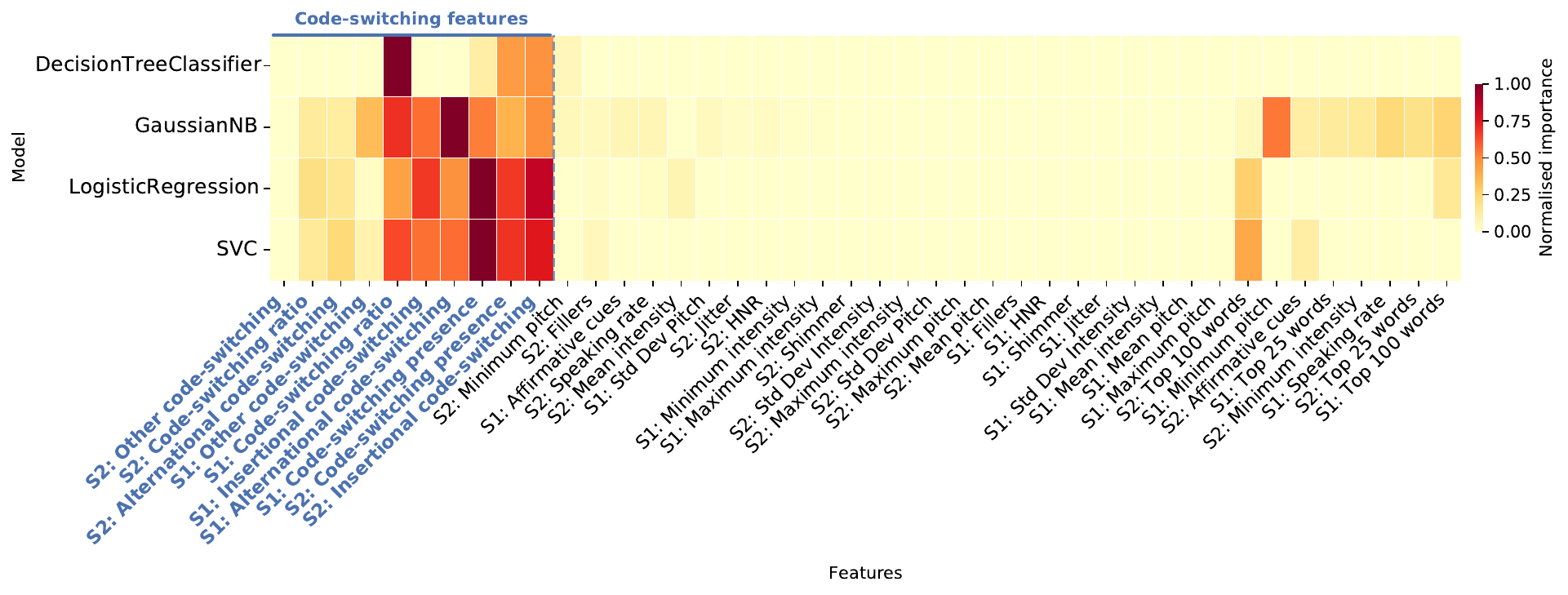}
    \caption{Feature importance values across the best-performing models from the secondary conversation-level analysis for MaSaC described in Appendix~\ref{sec:app_noisy_analysis}. We normalize values per model using row-wise min.-max.\ so that models with different importance scales are comparable. Unlike the corresponding figures for our primary analyses, the entraining/non-entraining label underlying this task is constructed directly from the CSW style features shown here, so the concentration of importance on these features reflects the label's construction rather than a positive, human-grounded finding; see Appendix~\ref{sec:app_noisy_analysis} for discussion.}  
    \label{fig:masac-feature-importances-noisy}
\end{figure*}

\begin{figure*}[!h]
\centering
\includegraphics[width=\linewidth]{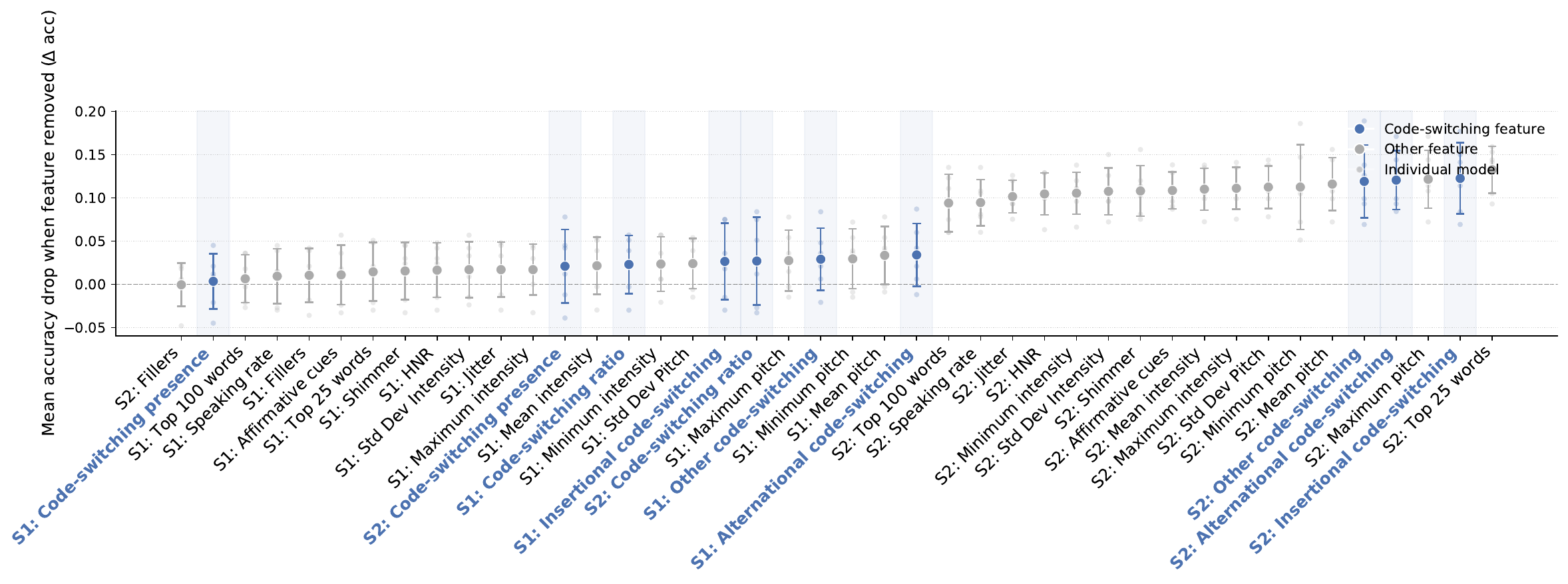}
    \caption{Aggregated feature ablation results for the secondary conversation-level analysis for MaSaC described in Appendix~\ref{sec:app_noisy_analysis}. Features are ordered left to right by increasing impact; $\Delta_{acc.}>0$ means accuracy \textit{drops} when the feature is removed, i.e., the feature is useful. As with the corresponding feature importance results (Figure~\ref{fig:masac-feature-importances-noisy}), the entraining/non-entraining label for this task is constructed directly from the CSW style features shown here; the resulting ablation pattern should not be interpreted as evidence that classifiers attend to human-relevant CSW style features when detecting entrainment. }
    \label{fig:masac-ablations-noisy}
\end{figure*}

\end{document}